%% file: paper_arxiv.tex
\documentclass{article}
\usepackage[nonatbib, preprint]{neurips_2026}
\usepackage{amsmath,amssymb,amsfonts}
\usepackage{graphicx}
\usepackage{booktabs}
\usepackage{hyperref}
\usepackage{url}
\usepackage{algorithm}
\usepackage{algpseudocode}
\usepackage{natbib}
\usepackage{microtype}
\usepackage{xcolor}
\usepackage{subcaption}
\usepackage{pifont}
\usepackage{listings}
\usepackage{booktabs}
\usepackage{colortbl}
\usepackage{multirow}
\graphicspath{{./figures/}}

\usepackage{listings}
\usepackage{xcolor}
\usepackage{tcolorbox}
\usepackage{booktabs}
\usepackage{titletoc}

\definecolor{varblue}{RGB}{24,95,165}
\definecolor{commentgray}{RGB}{120,120,120}
\definecolor{sectiongray}{RGB}{90,90,90}
\definecolor{rulegray}{RGB}{200,200,200}
\definecolor{createblue}{RGB}{24,95,165}
\definecolor{crossovergreen}{RGB}{29,158,117}
\definecolor{refinecoral}{RGB}{216,90,48}
\definecolor{mutateviolet}{RGB}{127,119,221}
\definecolor{codebg}{RGB}{250,250,249}
\definecolor{mutategold}{RGB}{180, 120, 0}
\lstdefinestyle{promptstyle}{
  basicstyle=\small\ttfamily,
  breaklines=true,
  frame=none,
  backgroundcolor=\color{codebg},
  xleftmargin=1em,
  xrightmargin=1em,
  aboveskip=0pt,
  belowskip=0pt,
  moredelim=[is][\color{varblue}\bfseries]{|[}{]|},
  moredelim=[is][\color{commentgray}\itshape]{|<}{>|},
  moredelim=[is][\color{sectiongray}\bfseries\small]{|{}{}|},
}

\title{RAISE: LLM-based Automated Heuristic Design with Robust Adversary Instance Search}

\author{
Fei Liu$^{1,2}$ \quad Alessio Figalli$^{2}$ \quad Patrick Owen$^{1}$ \quad Nicola Serra$^{1}$ \\
$^{1}$University of Zurich \quad  $^{2}$ETH Zurich 
}
\begin{document}

\maketitle

\begin{abstract}
Automated Heuristic Design (AHD) with Large Language Models (LLMs) has shown remarkable progress in discovering high-quality heuristics. However, existing LLM-based AHD methods optimize heuristics for a fixed training instance set and may fail catastrophically when deployed under real-world distributional shifts. We propose \underline{\textbf{R}}obust \underline{\textbf{A}}dversary \underline{\textbf{I}}nstance \underline{\textbf{Se}}arch (\textbf{RAISE}), a framework that integrates constrained worst-case instance search within a principled neighborhood of the training distribution into the LLM-based evolutionary search loop. RAISE treats robust AHD as a constrained adversarial instance search problem: the outer loop evolves heuristics via LLM operators, while an LLM-free inner loop efficiently identifies hard instances within an $\varepsilon$-ball around the training instance set using a basis distribution parameterization with boundary projection. Comprehensive experiments on Online Bin Packing (OBP), Online Job Shop Scheduling (OJSP), and Online Vehicle Routing (OVRP) across five distribution families demonstrate that existing LLM-based AHD methods degrade by up to $19\times$ under distribution shift, while RAISE consistently maintains strong performance across all tested distributions and problem scales.

\end{abstract}

\section{Introduction}
\label{sec:intro}

Designing effective heuristic algorithms for optimization problems has traditionally required extensive domain expertise and manual engineering.
Automated Heuristic Design (AHD) seeks to reduce this burden by automatically discovering high-quality heuristic programs, enabling rapid adaptation to new problem domains without much human intervention~\citep{pillay2018hyper,stutzle2018automated,qu2020general,zhou2024collaboration,ma2025toward}.

The recent integration of Large Language Models (LLMs) into evolutionary search has dramatically advanced AHD~\citep{liu2024eoh,liu2023algorithm,van2024llamea,ye2024reevo,dat2025hsevo,zheng2025monte,hu2025partition, ma2026llamoco,shi2026generalizable,liu2026eohs,novikov2025alphaevolve,sun2026co}. For example, \citet{liu2024eoh} propose Evolution of Heuristics (EoH), which uses LLMs as evolutionary operators to iteratively generate, recombine, and refine heuristic functions for combinatorial optimization. \citet{ye2024reevo} extend this paradigm with ReEvo, incorporating reflective short-term and long-term memory mechanisms that guide the LLM toward progressively better heuristics. Additionally, MCTS-AHD~\cite{zheng2025monte} and PartEvo~\cite{hu2025partition} integrate Monte Carlo tree search and feature-assisted niche construction, respectively, to enhance the efficiency of evolutionary search.


\textbf{The distribution shift problem.}
Despite their strong nominal performance (performance on training instance set), existing LLM-based AHD methods share a critical limitation: they optimize heuristics for a fixed instance distribution encountered during training, producing highly specialized solutions that may \emph{fail catastrophically} when deployed under different conditions~\citep{wang2026rethinking,shi2026generalizable}. This robustness gap is particularly problematic in real-world deployment, where problem instances are drawn from unknown or shifting distributions~\citep{zhou2024collaboration,goh2025shield,wang2026rethinking}. 

Recent work has attempted to address this limitation. \citet{liu2026eohs} propose EoH-S, which learns a portfolio of diverse heuristics to improve cross-distribution performance. However, it relies on a predefined set of diverse training instances. \citet{shi2026generalizable} introduce MoH, a meta-optimization method for cross-distribution generalization, but still requires explicit multi-task training across manually-defined distributions. In both cases, the dependence on a predefined set of distributions limits practicality in dynamic or unpredictable environments. Furthermore, while \citet{karimi2025robust} and \citet{duan2025ealg} have studied adversarial instance generation, they do not impose constraints on distributions and rely on LLMs to generate instances, which lacks controllability and introduces additional cost.

\textbf{This work.}
We propose \textbf{RAISE} (\textbf{R}obust \textbf{A}dversary \textbf{I}nstance \textbf{S}earch for LLM-based Heuristic Design), which incorporates adversarial instance search into the LLM evolutionary search loop. RAISE formulates robust AHD as a \textbf{constrained} minimax optimization problem to find the best robust heuristic $h^*$:
\[
  h^* = \arg\max_{h}\; \min_{s' \in \mathcal{B}_\varepsilon(S)}\; \mathrm{eval}(h,\, s'),
\]
where $\mathcal{B}_\varepsilon(S)$ denotes the ball of radius $\varepsilon$ around the nominal instance set $S$. The outer maximization is performed by the LLM evolutionary search, while the inner minimization is handled by an adversarial instance search that discovers the hardest instances within the uncertainty set.

\textbf{Contributions.} This paper makes the following contributions:
\begin{itemize}
  \item \textbf{Problem formulation.}
    We formalize an instance-level robust AHD objective as a constrained minimax problem over an $\varepsilon$-ball uncertainty set, providing an operationalizable approximation to Distributionally Robust Optimization (DRO) that does not require distributional assumptions, to the best of our knowledge, the first constrained instance-level formulation in the LLM-based AHD literature.

  \item \textbf{Adversarial instance search.}
    We propose RAISE, a bi-level evolutionary framework with an LLM-driven outer loop for heuristic evolution and an LLM-free inner loop for worst-case instance discovery. The inner loop employs a basis distribution parameterization with boundary projection to efficiently explore the uncertainty set without additional LLM cost.

  \item \textbf{Empirical validation.}
    Experiments on three online combinatorial optimization tasks---OBP, OJSP, and OVRP---across five distribution families and 95 datasets with different distributions show that RAISE significantly improves the performance on shifted distributions among all learned methods, while existing LLM-based AHD methods degrade by up to $19\times$ under distribution shift and RAISE remains competitive on nominal instances.
\end{itemize}


\section{Methodology}
\label{sec:method}

\subsection{Problem Formulation: Robust Automated Heuristic Design (RAHD)}

Let $\mathcal{H}$ denote the space of heuristics. For a given design task, each heuristic
$h \in \mathcal{H}$ is evaluated on an instance $s$ via a black-box oracle
$\mathrm{eval}(h, s) \in \mathbb{R}$. Without loss of generality, we define eval(h, s) so that larger values indicate better performance. For minimization objectives in our experiments, we negate the raw metric before evaluation.
Standard AHD optimizes
\[
  h^* = \arg\max_{h \in \mathcal{H}}\, \mathbb{E}_{s \sim P_0}[\mathrm{eval}(h, s)]
\]
under a fixed nominal distribution $P_0$ (usually the distribution is not available, so it is a training instance set). This objective is brittle under distributional
shift: a heuristic tuned to $P_0$ may degrade sharply when test instances are drawn from a
nearby but a different distribution \citep{wang2026rethinking,shi2026generalizable,liu2026eohs}.

To address this limitation, we adopt a distributionally robust formulation. Given a nominal
instance set $S = \{s_1, \ldots, s_m\} \sim P_0$ and a robustness radius $\varepsilon > 0$, we define the uncertainty set
\[
  \mathcal{B}_\varepsilon(P_0) = \bigl\{P : d(P, P_0) \leq \varepsilon\bigr\},
\]
where $d$ is a distance measure between distributions. In the distribution-level formulation, $d$ captures the discrepancy between $P$ and $P_0$ directly~\citep{rahimian2019distributionally}. There are different distance metrics (e.g., optimal transport metrics)~\cite{rahimian2019distributionally} can be employed. In practice, however, the true distribution $P_0$ is unavailable; only a finite set of nominal instances is observed. We therefore adopt an instance-level approximation,
defining the practical, robust AHD objective as
\begin{equation}
  h^* = \arg\max_{h \in \mathcal{H}}\;\min_{s' \in \mathcal{B}_\varepsilon(S)}\;
  \mathrm{eval}(h,\, s'),
  \label{eq:robust_ahd}
\end{equation}
where
\[
  \mathcal{B}_\varepsilon(S) = \bigl\{s : \exists\, s_i \in S,\; d(s, s_i) \leq \varepsilon\bigr\}
\]
is the instance-level uncertainty set,  where $d(s, s_i)$ is a distance metric between instances (e.g., any $\ell_p$ norm or 
problem-specific similarity measure). In this work, we simply adopt the normalized $\ell_1$ distance,
\[
  d(s_{a}, s_{b}) = \frac{1}{n}\sum_{j=1}^{n} \lvert s_{a,j} - s_{b,j} \rvert,
\]
where $n$ is the dimension of the instance feature vector.


\begin{figure}[t]
    \centering
    \includegraphics[width=\linewidth]{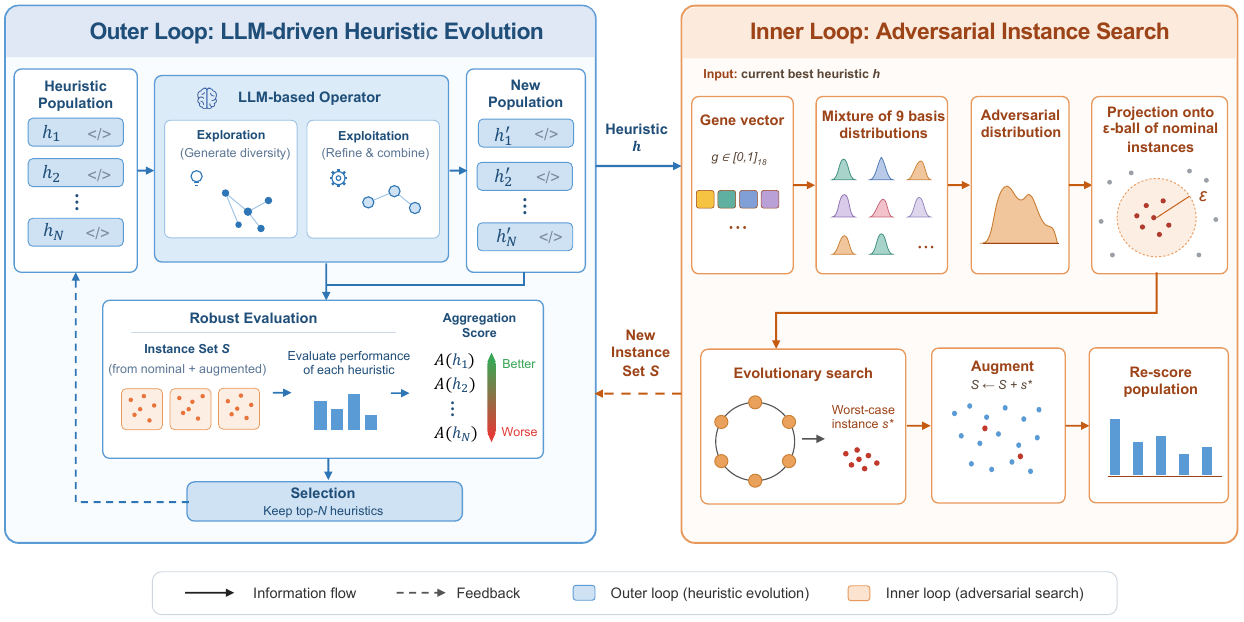}
    \caption{Overview of the RAISE framework: an LLM-driven outer evolutionary loop for heuristic design coupled with an LLM-free inner adversarial instance search that discovers worst-case instances within an $\varepsilon$-ball uncertainty set around the nominal training distribution.}
    \label{fig:placeholder}
\end{figure}

Equation~\eqref{eq:robust_ahd} is an instance-level approximation of distributionally robust optimization, rather than optimizing directly over all distributions in $\mathcal{B}_\varepsilon(P_0)$. 
We emphasize that this instance-level formulation is an engineering approximation that operationalizes robustness intuitions, rather than a formal solution to distribution-level DRO.

The robustness radius $\varepsilon$ continuously interpolates between two limiting regimes:
\begin{itemize}
    \item \textbf{When $\varepsilon \to 0$}, the uncertainty set $\mathcal{B}_\varepsilon(S)$ collapses to the nominal instance set $S$ itself, and the robust objective~\eqref{eq:robust_ahd} reduces to standard AHD---finding the heuristic with the best aggregated performance on the nominal instance set $S$, as in EoH~\citep{liu2024eoh} and ReEvo~\citep{ye2024reevo}. We note a subtle distinction: existing works typically aggregate performance via the mean over $S$, whereas our formulation uses a worst-case (min) aggregation; nonetheless, the instance set itself remains unchanged.

    \item \textbf{When $\varepsilon \to \infty$}, the uncertainty set expands to cover all
    feasible instances, and the objective becomes fully distribution-agnostic: finding the
    heuristic with the best worst-case performance across \emph{any} conceivable instance.
\end{itemize}

In practice, $\varepsilon$ is set to a finite intermediate value, providing a good balance of robustness performance. Larger values of $\varepsilon$ yield more conservative but shift-resilient heuristics, while smaller values recover the higher nominal performance of standard AHD.

\subsection{RAISE Framework Overview}

\begin{algorithm}[t]
\caption{RAISE: Robust Adversary Instance Search for LLM-Based Heuristic Design}
\label{alg:raise}
\begin{algorithmic}[1]
\Require Task description $\mathcal{T}$, nominal instances $S_0$, robustness radius
  $\varepsilon$, max samples $N_{\max}$, max pop size $\mathcal{P}_{\max}$, refresh interval $\tau$
\Ensure Best heuristic $h^*$
\State Initialize population $\mathcal{P} \leftarrow \emptyset$, instance set
  $\mathcal{S} \leftarrow \{S_0\}$
\State Sample $\mathcal{P}_{\max}$ heuristics, evaluate on
  $\mathcal{S}$, initialize $\mathcal{P}$
\While{$n_{\mathrm{samples}} < N_{\max}$}
  \For{each LLM operator} \Comment{Outer loop}
    \State Sample parent(s) via rank-weighted selection from $\mathcal{P}$
    \State Generate candidate $h$ via the selected LLM operator
    \State Evaluate $\hat{r}(h) \leftarrow \mathcal{A}\!\left(\bigl\{\mathrm{eval}(h,\, s'_k)\bigr\}_{k=1}^{K}\right)$
    \State Update $\mathcal{P}$ via survival selection
  \EndFor
  \If{generation $\bmod\, \tau = 0$} \Comment{Inner loop}
    \State Run inner adversarial search on current best $h \in \mathcal{P}$ and get worst instance $s^*$;
    \State Update $\mathcal{S} \leftarrow \mathcal{S} + s^*$
    \State Re-score all heuristics in $\mathcal{P}$ on updated $\mathcal{S}$
  \EndIf
\EndWhile
\State \Return $h^* \leftarrow \arg\max_{h \in \mathcal{P}} \hat{r}(h)$
\end{algorithmic}
\end{algorithm}

RAISE solves the minimax problem~\eqref{eq:robust_ahd} through a bi-level evolutionary search
with two coupled loops. The overall workflow is:

\begin{enumerate}
  \item Start from a nominal instance set $S$ and initialize a population of candidate
    heuristics.
  \item Evolve heuristics using LLMs in the \textbf{outer loop (LLM-based heuristic evolution)} under a fixed set of adversarial evaluation
    instances.
  \item Every $\tau$ generations, run the \textbf{inner loop (adversarial instance search)} to search for worst-case instances
    within the instance-level uncertainty set $\mathcal{B}_\varepsilon(S)$.
  \item Augment the instance set with the discovered hard instances, re-score the population,
    and continue the outer search.
\end{enumerate}

This alternating procedure approximates the minimax objective in~\eqref{eq:robust_ahd}: the
inner loop exposes current failure modes of the heuristic population, and the outer loop
adapts that population to those harder conditions. Algorithm~\ref{alg:raise} summarizes the
interaction between the two loops.


While formal convergence guarantees are not available, this alternating design has a clear
operational interpretation: the method repeatedly identifies weaknesses of the current
heuristic population and then selects against those weaknesses in the next outer-loop phase.

\subsection{Outer Loop: LLM-Based Evolutionary Heuristic Search}

The outer loop maintains a population $\mathcal{P}$ of heuristic programs and improves it
using LLM operators (e.g., synthesize a new heuristic from two high-performing heuristics or perturb one heuristic to explore its variants). The specific prompts are introduced in Appendix~\ref{app:method_detail}. 


After new heuristics are generated, each candidate $h$ is evaluated on the current adversarial instance set
$\mathcal{S} = \{s'_1, \ldots, s'_K\} \subseteq \mathcal{B}_\varepsilon(S)$ through the
robust score
\[
  \hat{r}(h) = \mathcal{A}\bigl(\mathrm{eval}(h, s'_1), \ldots, \mathrm{eval}(h, s'_K)\bigr),
\]
where $\mathcal{A}$ is an aggregation function, $K$ is the number of instances in current instance set.
This finite-sample score serves as a practical surrogate for the worst-case objective
in~\eqref{eq:robust_ahd}: as the inner loop enlarges $\mathcal{S}$ with progressively
harder instances, high-scoring heuristics are precisely those that remain robust across
the most challenging distributional shifts encountered so far. After evaluation, candidate heuristics
are admitted into the population $\mathcal{P}$ via survival selection, retaining only the
top-$N$ heuristics.


In principle, the aggregation A should be the minimum, in direct correspondence with the worst-case objective in ~\eqref{eq:robust_ahd}. 

In practice, directly optimizing the minimum over a finite and evolving adversarial set can produce unstable selection dynamics dominated by individual outlier instances. We therefore use mean aggregation as a more stable surrogate objective during search while retaining adversarial instance augmentation to encourage robustness.


\subsection{Inner Loop: LLM-Free Adversarial Instance Search}
\label{sec:inner_search}

Every $\tau$ outer-loop generations, RAISE runs an \emph{LLM-free} evolutionary search to
find worst-case instances within $\mathcal{B}_\varepsilon(S)$ for the current best heuristic
$h \in \mathcal{P}$. Each candidate adversarial instance is parameterized by an
18-dimensional gene vector $\mathbf{g} \in [0,1]^{18}$.

\textbf{Basis distribution encoding.}
The distribution of a candidate instance $s'$ is constructed as a mixture of nine basis distributions centered on the nominal:
\begin{equation}
  p_{\mathrm{adv}} = \mathrm{clip}\!\left(\bar{p} + \sum_{i=1}^{9} w_i \cdot
  \phi_i(\mathbf{g}),\; 0, 1\right),
  \label{eq:adv_dist}
\end{equation}
where $\bar{p}$ is the mean nominal distribution, $w_i = g_i - 0.5$ are signed mixture weights, and $\phi_i$ are nine parametric basis distributions: Uniform, Small, Large, Center, Bimodal, Gaussian, Periodic, Poisson-like, and Peak (Appendix~\ref{app:method:inner_search} lists the detailed settings). The 18 genes split into two groups: 9 genes define the mixture weights and the remaining 9 control the shape hyperparameters and support length. This parameterization is expressive enough to represent diverse distributional shifts while remaining low-dimensional enough for efficient inner-loop search. Moreover, it is not anchored to any predefined distribution family. 

\textbf{Epsilon-ball projection.}
After constructing $p_{\mathrm{adv}}$ via~\eqref{eq:adv_dist}, we project it onto the
$\varepsilon$-ball of the nearest nominal instance to enforce feasibility with respect to
$\mathcal{B}_\varepsilon(S)$:
\[
  p_{\mathrm{proj}} = b + \frac{\varepsilon}{\max(\varepsilon,\, d(p_{\mathrm{adv}}, b))}
  \cdot (p_{\mathrm{adv}} - b),
\]
where $b$ is the closest nominal instance and $d(\cdot,\cdot)$ is the mean absolute difference defined in Section~\ref{sec:method}. When $d(p_{\mathrm{adv}}, b)$ is larger than $\varepsilon$, this projection places candidates near the boundary of $\mathcal{B}_\varepsilon(S)$.

\textbf{Inner evolutionary algorithm.}
The inner search applies uniform crossover and Gaussian mutation over $G_{\mathrm{in}}$ generations on a population of $P_{\mathrm{in}}$ candidate gene vectors, with each vector evaluated by its ability to degrade the current best heuristic $h$. The worst-scoring feasible instances $s^*$, minimizing $\mathrm{eval}(h, s')$ within $\mathcal{B}_\varepsilon(S)$, is appended to instance set $\mathcal{S}$ for the outer loop. For example, if the current best heuristic is specialized to small-item instances on OBP, the inner loop may discover a large-item-heavy distribution. The outer loop then reevaluates all heuristics on the new worst instance and favors candidates that remain effective under the new set $\mathcal{S}$. 



\section{Experimental Studies}
\label{sec:experiments}


\subsection{Experimental Setup}

\paragraph{Tasks:} \textbf{1) Online Bin Packing (OBP):} $n$ items arrive sequentially and are  assigned to bins with a capacity $C$. Waste ratio $= (\text{bins used} - \text{lower bound}) / \text{lower bound}$ (lower is better) is used as the objective. Trained on 5 Weibull instances~\citep{romera2024mathematical,liu2024eoh} and Tested with $C \in \{100, 200, 300, 400\}$ and $n \in \{1000, 5000, 10000\}$. Each size combination has 5 different distributions (Uniform, Normal, Lognormal, Exponential, and Triangular) of items sizes, which results in $4 \times 3 \times 5  = 60 $ datasets for OBP. \textbf{2) Online Job Shop Scheduling (JSP):} Jobs with stochastic processing times are assigned to machines online. The objective is to minimize normalized makespan. Trained on 5 uniform distribution instances with 10 machines ($m$) and 20 jobs ($j$). Tested on problem sizes: $m \in \{10,20\}$ and $j \in \{20,50\}$. With 5 distributions for each combination, there is in total 20 datasets. \textbf{3) Online Vehicle Routing (VRP):} Customers with stochastic demands arrive online. The objective is to minimize normalized route length ratio. Trained on 5 instances with 10 vehicles ($v$) and 50 customers. Tested on problem sizes: $v \in \{5,10,15\}$. With 5 distribution for each combination, there is in total 15 datasets. The details of the three tasks and the five different distributions families are introduced in Appendix~\ref{app:exp_detail}. 

\vspace{-5pt}
\paragraph{Baselines.}
\emph{Classical:} BestFit and FirstFit (OBP); SPT, MinSlack, EDD (JSP); Nearest Feasible Insertion, Slack-Preserving Insertion, and Urgency-Weighted Insertion (VRP). \emph{LLM-based AHD:} \textbf{EoH}~\citep{liu2024eoh}, \textbf{ReEvo}~\citep{ye2024reevo}, and \textbf{PartEvo}~\citep{hu2025partition}, each optimised on the nominal training set. \emph{Robustness-aware LLM-based AHD:} \textbf{EoH-S}~\citep{liu2026eohs} (top heuristic trained on 128 diverse instances) and \textbf{MoH}~\citep{shi2026generalizable} (evaluated with the heuristic reported in the original paper).

 \vspace{-5pt}
\paragraph{Settings.}
All LLM-driven methods use 1{,}000 samples and a population size of 10, with GPT-5-mini as the backbone. All methods train on 5 nominal instances except EoH-S, which uses 128 diverse instances as in the original paper. For RAISE, the refresh interval is $\tau = 5$, with inner population $P_{\mathrm{in}} = 8$ and $G_{\mathrm{in}} = 4$ inner generations. We ablate four values $\varepsilon \in \{0.001, 0.002, 0.005, 0.010\}$ on OBP and use $\varepsilon = 0.002$ for OJSP and OVRP. All hyperparameter and prompt details are in
Appendix~\ref{app:method_detail}.

\begin{figure}[t]
  \centering
  \includegraphics[width=1.0\linewidth]{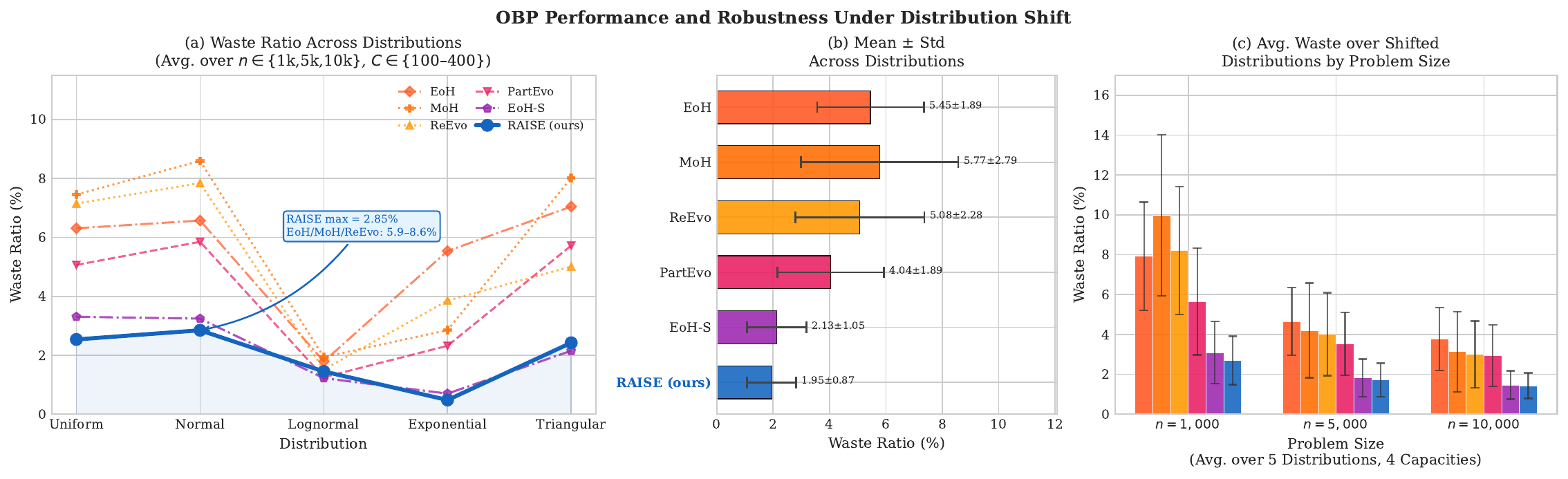}
  \caption{a) OBP waste ratio (the lower the better) across five distributions. b) Average results averaged on all 60 intance datasets. c) Averaged results across different sizes.}
  \label{fig:obp_distribution}
\end{figure}

\begin{table}[t]
\centering
\caption{OBP waste ratio ($\%$) on $n{=}5000$, $C{=}200$ instances. Best per distribution in \textbf{bold}. \underline{underline}: second best. Lower is better.}
\label{tab:obp_main}
\small
\begin{tabular}{lcccccc}
\toprule
\textbf{Method}  & \textbf{Uniform} & \textbf{Normal} &
  \textbf{Lognormal} & \textbf{Exponential} & \textbf{Triangular} &
  \textbf{Avg.} \\
\midrule
BestFit  & \underline{1.717} & \underline{2.803} & 1.914 & \underline{0.300} & 3.006 & 1.948 \\
FirstFit  & 2.928 & 4.096 & 2.049 & 0.450 & 3.696 & 2.644 \\
\midrule
EoH \cite{liu2024eoh} & 5.182 & 6.308 & 1.321 & 5.303 & 5.765 & 4.776 \\
ReEvo \citep{ye2024reevo} & 5.466 & 6.539 & \textbf{0.714} & 3.011 & 3.218 & 3.790 \\
PartEvo \citep{hu2025partition} & 3.853 & 5.062 & \underline{0.835} & 2.236 & 4.619 & 3.321 \\
EoH-S \citep{liu2026eohs}  & 2.318 & 3.010 & 1.334 & 0.436 & 2.043 & \underline{1.828} \\
MoH \citep{shi2026generalizable} & 5.277 & 6.531 & 0.917 & 1.443 & 5.113 & 3.856 \\
\midrule
RAISE$^{\varepsilon=0.001}$  & 3.814 & 4.225 & 0.822 & 3.435 & 1.997 & 2.859 \\
RAISE$^{\varepsilon=0.002}$  & 1.954 & 2.914 & 1.253 & 0.314 & \textbf{1.919} & \textbf{1.671} \\
RAISE$^{\varepsilon=0.005}$  & 2.785 & 3.402 & 1.011 & \textbf{0.273} & 2.636 & 2.021 \\
RAISE$^{\varepsilon=0.010}$  & \textbf{1.670} & \textbf{2.763} & 2.062 & 0.354 & 2.803 & 1.930 \\
\bottomrule
\end{tabular}
\end{table}

\begin{table*}[t]
\centering
\small
\setlength{\tabcolsep}{4pt}
\caption{Average waste ratio (\%) over five out-of-distribution test sets (Uniform, Normal, Lognormal, Exponential, Triangular) on Online Bin Packing across twelve size configurations ($n \in \{1\text{k},5\text{k},10\text{k}\}$, $C \in \{100,200,300,400\}$). \textbf{Bold}: best per column;  \underline{underline}: second best. Lower is better.}
\label{tab:obp_size_results}
\begin{tabular}{l|rrr|rrr|rrr|rrr}
\toprule
\multirow{2}{*}{\textbf{Method}} & \multicolumn{3}{c|}{$C=100$} & \multicolumn{3}{c|}{$C=200$} & \multicolumn{3}{c|}{$C=300$} & \multicolumn{3}{c}{$C=400$} \\
\cmidrule(lr){2-4}\cmidrule(lr){5-7}\cmidrule(lr){8-10}\cmidrule(lr){11-13}
 & $1\text{k}$ & $5\text{k}$ & $10\text{k}$ & $1\text{k}$ & $5\text{k}$ & $10\text{k}$ & $1\text{k}$ & $5\text{k}$ & $10\text{k}$ & $1\text{k}$ & $5\text{k}$ & $10\text{k}$ \\
\midrule
\textsc{BestFit} & \textbf{2.645} & 1.882 & 1.608 & \textbf{2.636} & 1.948 & 1.675 & \underline{2.656} & 1.984 & 1.725 & \textbf{2.644} & 2.006 & 1.732 \\
\textsc{FirstFit} & 3.570 & 2.527 & 2.147 & 3.629 & 2.644 & 2.287 & 3.633 & 2.691 & 2.341 & 3.647 & 2.714 & 2.367 \\
\midrule
\textsc{EoH} & 6.272 & 3.799 & 3.037 & 8.137 & 4.776 & 3.829 & 8.511 & 4.957 & 4.092 & 8.761 & 5.093 & 4.158 \\
\textsc{ReEvo} & 5.457 & 2.591 & 1.827 & 7.880 & 3.790 & 2.878 & 9.322 & 4.630 & 3.486 & 10.187 & 5.053 & 3.801 \\
\textsc{PartEvo} & 4.921 & 2.453 & 2.155 & 5.093 & 3.321 & 2.885 & 6.001 & 3.922 & 3.208 & 6.613 & 4.434 & 3.527 \\
\textsc{EoH-S} & 2.766 & \underline{1.741} & \underline{1.428} & 2.969 & \underline{1.828} & \underline{1.513} & 3.091 & \underline{1.865} & \underline{1.531} & 3.203 & \underline{1.899} & \textbf{1.563} \\
\textsc{MoH} & 6.217 & 2.632 & 1.937 & 9.310 & 3.856 & 2.867 & 11.355 & 4.685 & 3.550 & 13.016 & 5.640 & 4.199 \\
\midrule
\textsc{RAISE} (ours) & \underline{2.742} & \textbf{1.641} & \textbf{1.307} & \underline{2.685} & \textbf{1.671} & \textbf{1.377} & \textbf{2.629} & \textbf{1.731} & \textbf{1.478} & \underline{2.735} & \textbf{1.841} & \underline{1.566} \\
\bottomrule
\end{tabular}
\end{table*}
\subsection{Results Across Different Distributions}

Figure~\ref{fig:obp_distribution} summarizes the results on OBP.
Table~\ref{tab:obp_main} and Table~\ref{tab:obp_size_results} report average results over different distribution families and all twelve size configurations ($n \in \{1\mathrm{k}, 5\mathrm{k}, 10\mathrm{k}\}$, $C \in \{100, 200, 300, 400\}$).

\paragraph{LLM-based methods collapse under distribution shift.}
EoH reaches a waste ratio of $5.182\%$ on the Uniform out-of-distribution (OOD) setting and $6.308\%$ on Normal, while ReEvo reaches $5.466\%$ and $6.539\%$ respectively---both far above classical BestFit
($1.717\%$ / $2.803\%$). MoH, despite meta-level training, similarly collapses to $5.277\%$
on Uniform and $6.531\%$ on Normal, because it still relies on explicit multi-task training
across predefined distributions. These results confirm that heuristics optimised on a fixed
nominal distribution are highly specialised and brittle: even modest distributional shifts
cause severe performance collapse.

\paragraph{RAISE achieves consistent OOD robustness.}
RAISE$^{\varepsilon=0.002}$ attains an average shifted waste of $1.671\%$, outperforming all
baselines including the most competitive robustness-aware method EoH-S ($1.828\%$).
The gap is most pronounced on the Exponential distribution, where
RAISE$^{\varepsilon=0.005}$ achieves $0.273\%$ versus EoH's $5.303\%$, a $19.4\times$ improvement. Crucially, EoH-S trains on 128 diverse instances, which is substantially more data than RAISE's 5 nominal instances and is often impractical in real deployments where diverse training instances may be unavailable.

\paragraph{Robustness generalises across problem scales.}
Table~\ref{tab:obp_size_results} shows that RAISE's advantage is consistent across all twelve
size configurations. Among all learned methods, RAISE achieves the best average OOD
performance on 8 of 12 configurations and is the second best on the remaining four. Classical BestFit remains strong at $n = 1\text{k}$ owing to the lower
difficulty of small-instance packing, but RAISE decisively surpasses it at $n \geq 5\text{k}$.
These trends indicate that the robustness gains scale with instance size, aligning with the
expectation that adversarial instance search becomes more informative as problem scale grows.

\subsection{Results on Online JSP and VRP}

Table~\ref{tab:jsp_vrp_combined} reports OOD performance on JSP and VRP. RAISE achieves the lowest normalised makespan on both machine scales ($1.2266$ at 10-machine, $1.2390$ at 20-machine), outperforming EoH ($1.2284$ / $1.2539$) and ReEvo ($1.2422$ / $1.2500$). The advantage widens at the 20-machine scale ($\Delta = 0.011$ over EoH), where the richer combinatorial structure amplifies the benefit of adversarial worst-case exposure during search. All LLM-based methods substantially outperform classical scheduling rules, confirming the overall effectiveness of LLM-driven heuristic design.

RAISE attains the best normalised route-length ratio on the 5-vehicle ($0.8931$) and
10-vehicle ($0.9263$) configurations, outperforming EoH ($0.9017$ / $0.9273$) and ReEvo
($0.8966$ / $0.9287$). On the 15-vehicle configuration, RAISE ($0.9570$) is marginally
behind EoH ($0.9561$) while outperforming all classical insertion heuristics by a clear
margin. Performance differences among LLM-based methods are smaller on OVRP than on OBP,
suggesting that OVRP's combinatorial structure provides some inherent distributional diversity
across the tested OOD families. Nonetheless, RAISE consistently matches or exceeds competing
methods across all configurations, demonstrating that its robustness-aware search
generalises to structurally different online combinatorial problems beyond bin packing.

\begin{table*}[t]
\centering
\caption{%
  Performance on \textbf{online JSP} (normalized makespan, $\downarrow$) and
  \textbf{online VRP} (route-length ratio $\downarrow$).
  OJSP columns report the mean over 10 configurations
  (5 processing-time distributions $\times$ 2 job counts: 20j/50j);
  VRP columns report the mean over 5 demand distributions.
  Classical baselines are task-specific (\emph{---} = not applicable).
  \textbf{Bold}: best per column; \underline{underline}: second best.%
}
\label{tab:jsp_vrp_combined}
\small
\setlength{\tabcolsep}{7pt}
\begin{tabular}{l cc ccc}
\toprule
& \multicolumn{2}{c}{\textbf{Online JSP} (makespan $\downarrow$)}
& \multicolumn{3}{c}{\textbf{Online VRP} (route length, $\downarrow$)} \\
\cmidrule(lr){2-3}\cmidrule(lr){4-6}
\textbf{Method}
  & \textbf{10-machine} & \textbf{20-machine}
  & \textbf{5-vehicle} & \textbf{10-vehicle} & \textbf{15-vehicle} \\
\midrule
SPT              & 1.3561 & 1.3956 & ---    & ---    & ---    \\
MinSlack         & 1.3812 & 1.4346 & ---    & ---    & ---    \\
EDD              & 1.4233 & 1.4886 & ---    & ---    & ---    \\
NF-Insert        & ---    & ---    & 0.9347 & 0.9747 & 1.0028 \\
SP-Insert        & ---    & ---    & 0.8971 & 0.9360 & 0.9665 \\
UW-Insert        & ---    & ---    & 0.9000 & 0.9345 & 0.9670 \\
\midrule
EoH              & \underline{1.2284} & 1.2539 & 0.9017 & \underline{0.9273} & \textbf{0.9561} \\
ReEvo            & 1.2422 & \underline{1.2500} & \underline{0.8966} & 0.9287 & 0.9613 \\
\midrule
RAISE (ours)     & \textbf{1.2266} & \textbf{1.2390} & \textbf{0.8931} & \textbf{0.9263} & \underline{0.9570} \\
\bottomrule
\end{tabular}
\end{table*}

\subsection{Convergence process}

\begin{figure}[t]
    \centering
    \includegraphics[width=\linewidth]{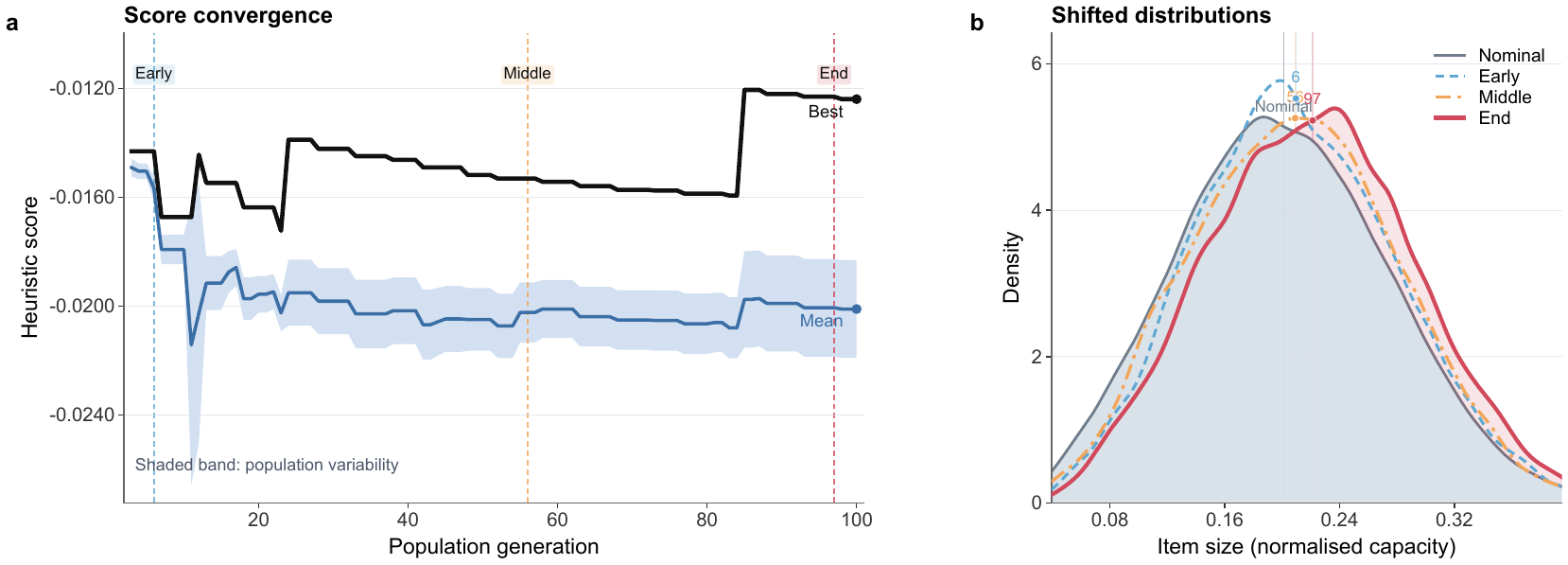}
    \caption{Robust Adversary Search Convergence in Online Bin Packing: \textbf{Left}, Population Score Evolution Across Generations; \textbf{Right}, Adversarial Distribution Shift Over Search Stages}
    \label{fig:convergence}
\end{figure}

Figure~\ref{fig:convergence} illustrates the characteristic convergence dynamics of RAISE on OBP under iterative robust adversary search. It shows the heuristic score (which should be larger the better, as a negative sign is used for illustration) as a function of the number of populations. The left panel presents the evolution of both the best heuristic score and the population mean over generations, revealing clear stage-wise behavior. Following an initial exploration phase, the population rapidly adapts to increasingly challenging evaluation instances, indicating progressive consolidation of robust design principles rather than isolated lucky discoveries. The ramp-up stage corresponds to the emergence of a new robust heuristic, while the decreasing stage reflects the exploration of new worst-case instances, during which the current heuristic score declines.

The right panel further reveals that this convergence is accompanied by a systematic shift in the adversarial item-size distribution away from the nominal regime toward harder distributions that expose the weakness of conventional packing rules. Taken together, these trends show that RAISE does not simply optimize for nominal performance, but converges by continually identifying, attacking, and repairing distribution-sensitive failure modes, ultimately yielding heuristics that remain effective under shifted distributions.

\section{Ablation Study}

We ablate four design choices of RAISE by independently removing each from the full system: (i)~\textbf{w/o Robust Search} removes the adversarial inner loop entirely, reducing RAISE to standard AHD; (ii)~\textbf{w/o Distribution Constraint} removes the $\varepsilon$-ball constraint-checking but keep the adversary search, which is similar to \cite{karimi2025robust} and \cite{duan2025ealg}; (iii)~\textbf{w/o Base Distributions} replaces the nine-basis parameterisation with flat-vector crossover and Gaussian mutation; (iv)~\textbf{w/o $\varepsilon$-Mapping} retains the nine-basis parameterisation but skips the boundary projection.

\begin{table}[t]
\centering
\caption{Ablation study on the Online Bin Packing (OBP) problem. We report the shifted waste ratio~(\%) under five out-of-distribution test settings, averaged over all problem sizes ($n \!\in\! \{1000, 5000, 10000\}$) and bin capacities ($C \!\in\! \{100, 200, 300, 400\}$). Lower is better; \textbf{bold} marks the best result per column.}
\label{tab:ablation_obp}
\setlength{\tabcolsep}{6pt}
\begin{tabular}{lcccccc}
\toprule
\multirow{2}{*}{\textbf{Method}}
  & \multicolumn{5}{c}{\textbf{Waste Ratio (\%) by Distribution}}
  & \multirow{2}{*}{\textbf{Avg.}} \\
\cmidrule(lr){2-6}
  & Uniform & Normal & Lognormal & Exponential & Triangular & \\
  \midrule
\addlinespace[2pt]
w/o Robust Search
  & 4.94 & 5.32 & 1.40
  & 1.21 & 5.11
  & 3.60  \\
w/o Dis. Constraints
  & 2.84 & 2.93 & \textbf{1.17}
  & 0.48 & 2.60
  & 2.01  \\
w/o Base Distributions
  & 4.49 & 4.84 & 2.77
  & 2.53 & 5.28
  & 3.98  \\
w/o $\varepsilon$-Mapping
  & 7.16 & 8.40 & 1.67
  & 2.33 & 10.23
  & 5.96  \\
\midrule
RAISE (full)
  & \textbf{2.54} & \textbf{2.85} & 1.45
  & \textbf{0.48} & \textbf{2.43}
  & \textbf{1.95}  \\
\bottomrule
\end{tabular}
\end{table}

Table~\ref{tab:ablation_obp} shows that removing any component consistently degrades OOD
robustness. Three findings stand out. (i) \textbf{The $\varepsilon$-mapping is the most critical component}, its removal raising the average waste ratio by $+4.01$ ($+205\%$) and causing catastrophic degradation on Triangular ($2.43\% \to 10.23\%$). Without boundary projection, the inner loop identifies adversarial instances far outside the nominal regime, causing the outer loop to over-optimise for distributional extremes at the expense of typical cases. (ii) \textbf{The nine-basis parameterisation provides essential distributional diversity.} Replacing it with flat-vector mutation increases average waste by $+2.03$ ($+104\%$), as plain perturbations fail to expose structurally different distributions such as skewed or bimodal item-size regimes. (iii) \textbf{The distribution constraint adds stability.} Removing the $\varepsilon$-ball projection while retaining the basis parameterisation yields $2.01\%$ average waste, marginally above the full model ($1.95\%$), confirming that constraining divergence from the nominal set prevents the search from concentrating on deployment-unlikely extremes. (iv) \textbf{The robust search procedure is indispensable}: ablating it entirely degrades average performance by $+1.65$ ($+85\%$), confirming that adversarial worst-case evaluation during search is necessary for consistent cross-distribution reliability.

\section{Conclusion}
\label{sec:conclusion}

We presented RAISE, a bi-level evolutionary framework that incorporates adversarial instance search under a constrained instance-level uncertainty set into LLM-based AHD. By formalizing robust AHD as a minimax problem over an instance-level $\varepsilon$-ball uncertainty set, RAISE decouples the robustness mechanism from any predefined distribution family: an LLM-driven outer loop evolves heuristics while an LLM-free inner loop continuously surfaces worst-case instances within a principled neighborhood of the training distribution, allowing RAISE to train on as few as five nominal instances yet generalize across substantially shifted distributions at test time. Comprehensive experiments on OBP, Online JSP, and Online VRP demonstrate that existing LLM-based AHD methods degrade by up to 19× under distribution shift, while RAISE consistently achieves the best average out-of-distribution performance among all learned methods across 95 test datasets.
\paragraph{Limitations and future work.} RAISE currently models single-dimensional distributional shifts and is evaluated only on online combinatorial optimization; extending to multi-dimensional correlated shifts, offline and mixed settings, and broader problem classes are important next steps.

\newpage

\bibliographystyle{plainnat}
\bibliography{references}

\appendix

\newpage
\section*{\centering \LARGE Appendix}
\addcontentsline{toc}{section}{Supplementary Material} 

\vspace*{10pt} 

\section*{Table of Contents}
\vspace*{-5pt} 

\startcontents[sections]

\printcontents[sections]{l}{1}{\setcounter{tocdepth}{2}}

\input{./sections/appendix_related_work}
\input{./sections/appendix_method_detail}
\input{./sections/appendix_experiment_detail}

\input{./sections/appendix_results_detail_new}
\input{./sections/appendix_heuristic_code}

\input{./sections/appendix_cost_analysis}

\input{./sections/appendix_license_and_llmusage}
\end{document}

%% file: sections/appendix_related_work.tex
\section{Related Work}
\label{app:related_work}

\subsection{LLM-Based Automated Heuristic Design}
\label{app:rw:llm_ahd}

The explosion of capable LLMs has unlocked a new paradigm for AHD in which the LLM serves as a black-box mutation and crossover operator over algorithm space. EoH \citep{liu2024eoh} is one of the preliminary works on LLM-based AHD. It maintains a population of heuristic Python functions, each scored on a fixed set of problem instances, and uses four LLM prompt types to create, crossover, and mutate programs. EoH achieves state-of-the-art performance on online bin packing and traveling salesman problems under nominal distribution conditions. ReEvo \citep{ye2024reevo} extends EoH with a dual-memory architecture: a short-term memory of recent high-scoring programs and a long-term memory of reflective insights generated by the LLM analyzing performance gaps. ReEvo shows that guided memory injection substantially improves sample efficiency, achieving competitive results with fewer LLM calls. HSEvo \citep{dat2025hsevo} proposes a harmony search framework to balance exploitation and exploration. MCTS-AHD \citep{zheng2025monte} employs Monte Carlo tree search to organize heuristics in a tree structure, enabling more effective exploration of the search space. PartEvo \citep{hu2025partition} incorporates feature-assisted niche construction within abstract search spaces, enabling the seamless integration of niche-based search strategies from evolutionary computation.

\paragraph{Generalizable heuristic design}
EoH-S \citep{liu2026eohs} introduces portfolio-based AHD: instead of seeking a single best heuristic, it discovers a diverse set of complementary heuristics with varying strengths. EoH-S is particularly relevant as a baseline in our distribution-shift experiments, as maintaining a diverse portfolio implicitly confers some cross-distribution robustness. MoH \citep{shi2026generalizable} leverages LLMs to iteratively refine a meta-optimizer that autonomously constructs diverse heuristic-optimizers through self-invocation, thereby eliminating reliance on a predefined evolutionary computation heuristic-optimizer. These constructed heuristic-optimizers subsequently evolve heuristics for downstream tasks, enabling broader heuristic exploration. MoH further employs a multi-task training scheme to promote generalization.

A key limitation shared by all of the above methods is that they optimize for a \emph{fixed} nominal instance distribution. While EoH-S and MoH propose different strategies to enhance generalization, both still rely on predefined instance sets and offer no guarantee of robust performance under unknown distributions. RAISE explicitly addresses this limitation via distributional robustness through adversarial instance search.

\paragraph{Adversarial instance search.}
\citet{karimi2025robust} employ an LLM to identify regions of the input space where a heuristic underperforms and to suggest targeted improvements. \citet{duan2025ealg} propose a co-evolutionary framework that jointly evolves both the heuristic algorithm and an instance generator using LLMs, with dynamically generated instances serving to enhance heuristic robustness throughout the evolutionary process.

While both approaches incorporate adversarial search into LLM-driven AHD, they impose no explicit constraints on the instance distribution and rely on LLMs to generate adversarial instances directly, which limits controllability and incurs additional inference cost. In contrast, RAISE employs a \emph{constrained} adversarial search that bounds the divergence between adversarial and nominal instance distributions, striking a principled balance between worst-case robustness and performance on the nominal distribution. Crucially, the worst-case instance search loop in RAISE is LLM-free, making it lightweight and ready to work with a broad range of LLM-driven AHD frameworks.

\subsection{Hyper-Heuristics and Automated Algorithm Configuration}
\label{app:rw:hyper}

\citet{burke2013hyper} surveys hyper-heuristics across two dimensions: (1) heuristic selection
vs.\ heuristic generation, and (2) constructive vs.\ perturbative. RAISE belongs to the
\emph{generative, constructive} quadrant---it synthesizes new heuristic functions from scratch
rather than selecting or composing existing low-level heuristics.

Traditional hyper-heuristics rely on hand-crafted low-level heuristic libraries and meta-level
selectors (e.g., simulated annealing, reinforcement learning) to adaptively combine them.
This approach is interpretable and efficient but fundamentally limited by the quality of the
low-level library.

Algorithm configuration methods such as SMAC \citep{hutter2011sequential}, ParamILS \citep{hutter2009paramils}, and irace \citep{lopez2016irace}
find optimal hyperparameter settings for fixed algorithm templates using model-based search.
They operate in continuous or mixed-integer parameter spaces, enabling principled exploration,
but they cannot \emph{design} new algorithmic logic.





\subsection{Adversarial Training and Minimax Learning}
\label{app:rw:adversarial}

DRO \citep{rahimian2019distributionally} is a classical framework in operations research and machine learning for decision-making under distributional uncertainty. The key idea is to replace the expected objective $\mathbb{E}_{P_0}[f(x)]$ with the worst-case expected objective $\sup_{P \in \mathcal{U}} \mathbb{E}_P[f(x)]$ over an ambiguity set $\mathcal{U}$.

Adversarial training \citep{goodfellow2014generative} and minimax learning originate in the GAN framework, where a generator and discriminator are co-trained via a minimax objective. The connection to distributional robustness was made explicit in subsequent work: training on adversarially perturbed data provably improves worst-case generalization under the assumed perturbation model.

Domain adaptation methods \citep{ben2010theory} study the theoretical conditions under which a model trained on a source distribution generalizes to a target distribution, providing H-divergence-based bounds on transfer error. These bounds motivate training procedures that minimize distribution shift, aligning with RAISE's adversarial instance generation objective.

Distributionally robust neural networks \citep{Sagawa2020Distributionally} apply group DRO to improve worst-case accuracy across subgroups in neural network training, demonstrating that explicit worst-case group weighting outperforms uniform ERM under distribution shift. RAISE is conceptually analogous but operates in the program-synthesis domain: instead of gradient-based adversarial perturbation of network weights, RAISE uses an evolutionary inner loop over a parameterized distribution space, since heuristic programs are non-differentiable.

The key distinction between RAISE and all gradient-based adversarial/DRO methods is the \emph{black-box} nature of heuristic evaluation: there is no gradient of the performance metric with respect to instance parameters, necessitating derivative-free optimization.

%% file: sections/appendix_method_detail.tex
\section{Detailed Method Description}
\label{app:method_detail}

This appendix provides comprehensive documentation of RAISE's components beyond what is possible
in the main paper. We describe each component in detail, provide complete pseudocode for
sub-routines, document all LLM prompt templates, give a full hyperparameter table, and discuss
implementation details and theoretical properties.

\subsection{Bi-Level Search Architecture}
\label{app:method:bilevel}

RAISE's bi-level structure separates two timescales: the \emph{outer loop} runs continuously (every generation), while the \emph{inner loop} fires periodically (every $\tau = 5$ generations). This separation is motivated by computational efficiency: the inner adversarial search involves re-evaluating an inner population of 8 candidate distributions over the current best heuristic; amortizing this cost over $\tau$ outer generations keeps the overhead below 20\% of total compute. 

At each inner loop invocation, the best heuristic $h^*_t$ from the current outer population is used as the fixed ``victim'' for the adversarial search. The search finds instances that minimize $\mathrm{eval}(h^*_t, s')$ subject to $d(s', s_i) \leq \varepsilon$ for some $s_i \in S$, where $d$ denotes the mean absolute difference defined in Section~\ref{sec:method}. After the inner search, the worst-case instance is appended to the current instance set $\mathcal{S}$, and the entire outer population is re-scored under the updated instances. This re-scoring ensures consistent ranking of all heuristics under the same adversarial context.

\subsection{Inner Adversarial Instance Search}
\label{app:method:inner_search}

The inner evolutionary algorithm is a compact evolutionary search. Its goal is to find, within the feasible uncertainty set $\mathcal{B}_\varepsilon(S)$, the instances that most severely degrade the performance of the current best heuristic $h^*$. 

Rather than searching directly in the space of problem instances---which is high-dimensional and unstructured---the inner loop operates on a compact 18-dimensional gene space that parameterizes a rich family of item-size distributions via basis function mixtures. Each gene vector is decoded into a candidate adversarial distribution, projected onto the boundary of the uncertainty set to ensure feasibility, and evaluated by running $h^*$ on a sampled instance. The population is then improved over $G_{\mathrm{in}}$ generations using selection, crossover, and mutation. Algorithm~\ref{app:alg:inner_search} provides the complete pseudocode; the remainder of this section describes each step in detail.

\begin{algorithm}[htbp]
\caption{Inner Adversarial Instance Search}
\label{app:alg:inner_search}
\begin{algorithmic}[1]
\Require Current best heuristic $h^*$, nominal instances $S = \{s_1,\ldots,s_k\}$,
  radius $\varepsilon$, inner population size $P_{\mathrm{in}}$,
  inner generations $G_{\mathrm{in}}$
\Ensure Worst-case feasible instance
\State Initialize gene population $\mathcal{G} = \{\mathbf{g}_1, \ldots, \mathbf{g}_{P_{\mathrm{in}}}\}$
  with $\mathbf{g}_i \sim \mathrm{Uniform}([0,1]^{18})$
\For{$g = 1$ to $G_{\mathrm{in}}$}
  \For{each $\mathbf{g}_i \in \mathcal{G}$}
    \State $p_{\mathrm{adv}} \leftarrow \mathrm{decode\_and\_mix}(\mathbf{g}_i)$
      \Comment{Eq.~\eqref{eq:adv_dist} in main paper}
    \State $b \leftarrow \arg\min_{s_j \in S} d(p_{\mathrm{adv}}, s_j)$
      \Comment{Nearest nominal instance}
    \State $p_{\mathrm{proj}} \leftarrow b + \frac{\varepsilon}{\max(\varepsilon,\, d(p_{\mathrm{adv}}, b))} \cdot (p_{\mathrm{adv}} - b)$
      \Comment{Epsilon-boundary projection}
    \State $s'_i \leftarrow \mathrm{sample\_instance}(p_{\mathrm{proj}})$
    \State $\mathrm{score}_i \leftarrow -\mathrm{eval}(h^*, s'_i)$
      \Comment{Adversarial score: lower eval = higher adversarial score}
  \EndFor
  \State Sort $\mathcal{G}$ by $\mathrm{score}$ descending; keep top $\lfloor P_{\mathrm{in}}/2 \rfloor$
  \State \textbf{Crossover}: for each pair of selected parents $(\mathbf{g}_i, \mathbf{g}_j)$,
    create offspring via uniform crossover with $p_{\mathrm{cross}} = 0.5$
  \State \textbf{Mutation}: apply Gaussian noise $\mathcal{N}(0, 0.12^2)$ to each gene with
    $p_{\mathrm{mut}} = 0.35$; clip to $[0,1]^{18}$
\EndFor
\State \Return The worst instance by adversarial score from final generation
\end{algorithmic}
\end{algorithm}

\textbf{Step 1: Population initialization (line 1).}
The inner search begins by randomly initializing $P_{\mathrm{in}} = 8$ gene vectors, each drawn independently from $\mathrm{Uniform}([0,1]^{18})$. This uniform initialization ensures broad coverage of the gene space at the start of the search, giving the evolutionary process diverse starting points without imposing any prior bias toward a particular distributional shift.

\textbf{Step 2: Decoding and distribution construction (line 4).}
Each gene vector $\mathbf{g}_i$ is decoded into a candidate adversarial item-size distribution $p_{\mathrm{adv}}$ via the basis-mixture model (Eq.~\eqref{eq:adv_dist} in the main paper). The first nine genes $g_{1:9}$ define signed mixture weights $w_i = g_i - 0.5 \in [-0.5, 0.5]$ over nine parametric basis distributions, while genes $g_{10:17}$ control the shape hyperparameters of those bases and $g_{18}$ adjusts the instance length within $\pm 10\%$ of the nominal. The resulting $p_{\mathrm{adv}}$ is a flexible, smooth distribution that can represent a wide range of shifts---heavy small-item tails, large-item concentrations, bimodal splits, periodic patterns, and others---while remaining low-dimensional enough for efficient evolutionary search. The nine basis distributions and their gene encodings are described in detail at the end of this section.

\textbf{Step 3: Nearest nominal instance and epsilon-ball projection (lines 5--6).}
The unconstrained distribution $p_{\mathrm{adv}}$ produced by decoding may fall anywhere in distribution space and need not satisfy the feasibility constraint $d(p_{\mathrm{adv}}, s_j) \leq \varepsilon$ for any $s_j \in S$. To enforce feasibility, we first identify the nearest nominal instance $b = \arg\min_{s_j \in S} d(p_{\mathrm{adv}}, s_j)$ under the mean absolute difference $d$, and then project $p_{\mathrm{adv}}$ radially onto the $\varepsilon$-boundary of the ball centered at $b$:
\[
  p_{\mathrm{proj}} = b + \frac{\varepsilon}{\max(\varepsilon,\, d(p_{\mathrm{adv}}, b))}
  \cdot (p_{\mathrm{adv}} - b).
\]
When $d(p_{\mathrm{adv}}, b) > \varepsilon$, this formula rescales the displacement vector $(p_{\mathrm{adv}} - b)$ so that the projected point lies exactly on the boundary at distance $\varepsilon$ from $b$; when $d(p_{\mathrm{adv}}, b) \leq \varepsilon$, the point is already feasible and is returned unchanged (since $\max(\varepsilon, d) = \varepsilon$ and the scaling factor equals 1). Projecting onto the boundary rather than the interior is deliberate: the hardest adversarial distributions concentrate near the boundary, where the maximum allowed shift from the nominal is achieved. Interior points represent smaller, less extreme shifts and are therefore typically less damaging to the heuristic.

\textbf{Step 4: Instance sampling and adversarial scoring (lines 7--8).}
From the projected distribution $p_{\mathrm{proj}}$, a concrete problem instance $s'_i$ is generated. The instance is then passed to the black-box evaluator to compute $\mathrm{eval}(h^*, s'_i)$, which measures how well $h^*$ performs on $s'_i$. The adversarial score is defined as the negation, $\mathrm{score}_i = -\mathrm{eval}(h^*, s'_i)$, so that maximizing the adversarial score is equivalent to finding instances on which $h^*$ performs worst. 

\textbf{Step 5: Selection (line 9).}
After scoring all $P_{\mathrm{in}}$ candidates in a generation, the population is ranked by adversarial score in descending order and the top half (i.e., $\lfloor P_{\mathrm{in}}/2 \rfloor = 4$ individuals) is retained as parents for the next generation. This truncation-selection scheme is computationally simple and well-suited to the short inner search horizon. The selected parents define the gene pool from which the next generation is produced via crossover and mutation.

\textbf{Step 6: Crossover (line 10).}
Each pair of selected parents $(\mathbf{g}_i, \mathbf{g}_j)$ produces one offspring via uniform crossover with probability $p_{\mathrm{cross}} = 0.5$: for each of the 18 gene positions independently, the offspring inherits the value from $\mathbf{g}_i$ with probability 0.5 and from $\mathbf{g}_j$ with probability 0.5. Uniform crossover is chosen over single-point or two-point crossover because it treats all gene positions symmetrically, which is appropriate here given that the 18 genes encode heterogeneous quantities (mixture weights, shape parameters, and a length parameter) with no meaningful positional ordering.

\textbf{Step 7: Mutation (line 11).}
Each gene of every offspring is independently perturbed by additive Gaussian noise $\mathcal{N}(0, 0.12^2)$ with probability $p_{\mathrm{mut}} = 0.35$, and the result is clipped to $[0,1]^{18}$ to maintain valid gene values. The mutation standard deviation $\sigma = 0.12$ is chosen to produce moderate perturbations relative to the unit gene range: on average, a mutated gene shifts by approximately $0.12 \times 0.35 \approx 0.04$ units, which translates to a meaningful but not catastrophic change in the decoded distribution. 

\textbf{Step 8: Return worst-case instances (line 13).}
After $G_{\mathrm{in}}$ generations, the worst candidate instance with the highest adversarial scores across the final population are returned to the outer loop. It is appended to the instance set $\mathcal{S}$, after which all heuristics in the outer population are re-scored on the updated $\mathcal{S}$ before evolution continues.

\textbf{Gene decoding detail.}
The 18-dimensional gene vector $\mathbf{g}$ is partitioned as:
\begin{itemize}
  \item $g_{1:9}$: basis function weights $w_i = g_i - 0.5 \in [-0.5, 0.5]$
  \item $g_{10:17}$: shape hyperparameters for basis distributions (e.g., power $p$ for
    Small/Large, mean/std for Gaussian, period for Periodic, peak position for Peak,
    center width for Center)
  \item $g_{18}$: length gene (fractional adjustment to item count within $\pm 10\%$ of nominal)
\end{itemize}

The nine basis distributions $\phi_i$ are:
\begin{enumerate}
  \item \textbf{Uniform}: $q \sim \mathrm{Uniform}(0,1)$
  \item \textbf{Small}: $q \sim \mathrm{Beta}(1, p)$ with shape $p = 1 + 4 \cdot g_{10}$
    (favors small items)
  \item \textbf{Large}: $q \sim \mathrm{Beta}(p, 1)$ with shape $p = 1 + 4 \cdot g_{11}$
    (favors large items)
  \item \textbf{Center}: $q \sim \mathrm{Triangular}(0, 0.5, 1)$ (concentrated around 0.5)
  \item \textbf{Bimodal}: mixture of $\mathrm{Uniform}(0, 0.3)$ and $\mathrm{Uniform}(0.7, 1)$
  \item \textbf{Gaussian}: $q \sim \mathrm{Normal}(\mu, \sigma^2)$ clipped to $[0,1]$,
    with $\mu = g_{12}$, $\sigma = 0.1 + 0.3 \cdot g_{13}$
  \item \textbf{Periodic}: sinusoidal weight with frequency controlled by $g_{14}$
  \item \textbf{Poisson-like}: discretized Poisson CDF with rate $\lambda = 1 + 9 \cdot g_{15}$
  \item \textbf{Peak}: spike at position $g_{16}$, spreading controlled by $g_{17}$
\end{enumerate}

\subsection{LLM Evolutionary Operators}
\label{app:method:operators}

The outer loop employs five LLM operators from \citep{liu2024eoh} with distinct roles, as described in
Section~\ref{sec:method}: \textsc{Create}, \textsc{E1}, \textsc{E2}, \textsc{M1}, and
\textsc{M2}. Table~\ref{app:tab:operators} summarizes their inputs, outputs, and
invocation characteristics. All operators require the model to first describe the
proposed heuristic in one sentence (enclosed in a boxed environment) before
implementing the Python function, ensuring interpretable outputs alongside executable code.

\begin{table}[htbp]
\centering
\caption{Summary of RAISE's five LLM evolutionary operators.}
\label{app:tab:operators}
\small
\begin{tabular}{@{}lllc@{}}
\toprule
\textbf{Operator} & \textbf{Inputs} & \textbf{Purpose} & \textbf{Parents} \\
\midrule
\textcolor{createblue}{\textsc{Create}}
  & Task description only
  & Generate novel heuristic from scratch & 0 \\[2pt]
\textcolor{crossovergreen}{\textsc{E1}}
  & $k$ parents + task
  & Generate heuristic with a totally different form from all parents & $k$ \\[2pt]
\textcolor{refinecoral}{\textsc{E2}}
  & $k$ parents + task
  & Extract backbone idea from parents; synthesize a new motivated variant & $k$ \\[2pt]
\textcolor{mutateviolet}{\textsc{M1}}
  & One parent + task
  & Produce a modified version of the parent heuristic & 1 \\[2pt]
\textcolor{mutategold}{\textsc{M2}}
  & One parent + task
  & Perturb numerical parameters of the parent's scoring function & 1 \\
\bottomrule
\end{tabular}
\end{table}

The operators divide into two families. The \emph{exploration} operators (\textsc{Create},
\textsc{E1}, \textsc{E2}) are designed to introduce structural novelty: \textsc{Create}
requires no parent at all; \textsc{E1} explicitly instructs the model to produce a form
\emph{totally different} from all provided parents; and \textsc{E2} first asks the model
to identify the common backbone idea across parents before synthesizing a new heuristic
motivated by---but structurally distinct from---that backbone.

The \emph{exploitation} operators (\textsc{M1}, \textsc{M2}) make incremental changes to a
single parent. \textsc{M1} allows free modification of the heuristic's logic, while
\textsc{M2} targets only the numerical parameter settings of the existing scoring function,
enabling fine-grained local search without altering algorithmic structure.

\begin{tcolorbox}[title=\textsc{Create} Prompt Template, colback=white, colframe=black, boxrule=0.4pt]
\{task\_description\}

\medskip
Create a new heuristic.
\begin{enumerate}
  \item First, describe your heuristic in one sentence inside boxed \{\}.
  \item Next, implement the following Python function:
        \{function\_signature\}
\end{enumerate}
Do not give additional explanations.
\end{tcolorbox}

\begin{tcolorbox}[title=\textsc{E1} Prompt Template, colback=white, colframe=black, boxrule=0.4pt]
\{task\_description\}

\medskip
I have \{k\} existing heuristics with their codes as follows: 

\{indivs\}

Please help me create a new heuristic that has a \textbf{totally different form}
from the given ones.
\begin{enumerate}
  \item First, describe your heuristic in one sentence. The description must be inside boxed \{\}.
  \item Next, implement the following Python function:
        \{function\_signature\}
\end{enumerate}
Do not give additional explanations.
\end{tcolorbox}

\begin{tcolorbox}[title=\textsc{E2} Prompt Template, colback=white, colframe=black, boxrule=0.4pt]
\{task\_description\}

\medskip
I have \{k\} existing heuristics with their codes as follows:

\{indivs\}

Please help me create a new heuristic that has a totally different form from the given
ones but can be \textbf{motivated} from them.
\begin{enumerate}
  \item Firstly, identify the common backbone idea in the provided heuristics.
  \item Secondly, based on the backbone idea describe your new heuristic in one sentence.
        The description must be inside boxed \{\}.
  \item Thirdly, implement the following Python function:
        \{function\_signature\}
\end{enumerate}
Do not give additional explanations.
\end{tcolorbox}

\begin{tcolorbox}[title=\textsc{M1} Prompt Template, colback=white, colframe=black, boxrule=0.4pt]
\{task\_description\}

\medskip
I have one heuristic with its code as follows:

\{indivs\}

Please create a new heuristic that can be a \textbf{modified version} of the given one.
\begin{enumerate}
  \item First, describe your heuristic in one sentence. The description must be inside boxed \{\}.
  \item Next, implement the following Python function:
        \{function\_signature\}
\end{enumerate}
Do not give additional explanations.
\end{tcolorbox}

\begin{tcolorbox}[title=\textsc{M2} Prompt Template, colback=white, colframe=black, boxrule=0.4pt]
\{task\_description\}

\medskip
I have one heuristic with its code as follows:

\{indivs\}

Please identify the main heuristic parameters and assist me in creating a new heuristic
that has \textbf{different parameter settings} of the score function provided.
\begin{enumerate}
  \item First, describe your heuristic in one sentence. The description must be inside boxed \{\}.
  \item Next, implement the following Python function:
        \{function\_signature\}
\end{enumerate}
Do not give additional explanations.
\end{tcolorbox}

%% file: sections/appendix_experiment_detail.tex
\section{Detailed Experimental Design and Settings}
\label{app:exp_detail}

This appendix provides complete documentation of the experimental setup, including
benchmark task definitions, dataset construction procedures with exact parameterizations
derived from the generation scripts, baseline algorithm configurations, RAISE hyperparameter
settings, and the full evaluation protocol.

\subsection{Benchmark Tasks and Instance Generation}
\label{app:exp:benchmarks}

\subsubsection{Online Bin Packing (OBP)}
\label{app:exp:bench_obp}

\textbf{Problem definition.}
In Online Bin Packing, items of integer size $a_i \in \{1, \ldots, C\}$ arrive one at a time
from an online sequence $\sigma = (a_1, a_2, \ldots, a_n)$.
Upon arrival, each item must be irrevocably assigned to an open bin (without exceeding the
bin capacity $C$) or to a new bin.
The objective is to minimize the total number of bins used.
Formally, let $B$ denote the number of bins used and
$L = \lceil \sum_i a_i / C \rceil$ the lower bound (optimal offline solution).
The \textbf{waste ratio} is:
\[
  \mathrm{WasteRatio} = \frac{B - L}{L}.
\]
Lower values indicate better performance; a waste ratio of 0 means the online algorithm
matches the offline lower bound.

\textbf{Training instances.}
RAISE's evolutionary search is seeded with a nominal OBP instance set comprising
$I = 5$ independent instances in Weibull distribution~\citep{romera2024mathematical,liu2024eoh}, each with $n = 5{,}000$ items and bin capacity $C = 200$.
These nominal instances constitute the initial adversarial instance set $\mathcal{S}_0$, which
RAISE augments dynamically during the search.
All LLM-based baselines (EoH, MoH, ReEvo, PartEvo, EoH-S) are trained on this same nominal
set to ensure a fair comparison.

\textbf{Test distribution parameterization.}
Testing evaluates cross-distribution generalization across five standard families.
Item sizes are drawn from the continuous distribution, rounded to the nearest integer
($\lfloor \cdot \rceil$), and clipped to the valid range $[1, C]$.
Table~\ref{app:tab:obp_dist_params} gives the exact parameterizations used in the
generation script.

\begin{table}[htbp]
\centering
\caption{OBP item-size distribution parameterizations (capacity $C$).
  All samples are rounded to integers and clipped to $[1, C]$.}
\label{app:tab:obp_dist_params}
\small
\setlength{\tabcolsep}{6pt}
\begin{tabular}{llll}
\toprule
\textbf{Distribution} & \textbf{Continuous draw} & \textbf{Parameters} & \textbf{Support} \\
\midrule
Uniform     & $a_i \sim \mathcal{U}(1,\,C)$                                      & ---                                        & $[1, C]$ \\
Normal      & $a_i \sim \mathcal{N}(\mu,\,\sigma^2)$                             & $\mu=0.5C,\;\sigma=0.2C$                  & clip to $[1, C]$ \\
Lognormal   & $a_i \sim \mathrm{LogNormal}(\mu_\ell,\,\sigma_\ell^2)$            & $\mu_\ell = \ln(0.25C),\;\sigma_\ell=0.6$ & clip to $[1, C]$ \\
Exponential & $a_i \sim \mathrm{Exp}(\lambda)+1$                                 & $\lambda = 0.3C$                           & clip to $[1, C]$ \\
Triangular  & $a_i \sim \mathrm{Tri}(\ell,\,m,\,r)$                             & $\ell=1,\;m=0.35C,\;r=C$                  & $[1, C]$ \\
\bottomrule
\end{tabular}
\end{table}

\textbf{Test suite structure.}
The OBP test suite spans all five distributions, three problem sizes ($n \in \{1{,}000,\;5{,}000,\;10{,}000\}$), and four bin capacities ($C \in \{100,\;200,\;300,\;400\}$), yielding $5 \times 3 \times 4 = 60$ dataset files, each containing $I = 5$ independently generated instances. The complete test suite thus covers \textbf{300 OBP test instances}. Table~\ref{app:tab:obp_test_suite} summarizes the combinatorial structure.

\begin{table}[htbp]
\centering
\caption{OBP test suite summary.
  Each (distribution, $n$, $C$) combination produces one dataset file with $I=5$ instances.}
\label{app:tab:obp_test_suite}
\small
\setlength{\tabcolsep}{6pt}
\begin{tabular}{lcccr}
\toprule
\textbf{Problem size $n$} & \textbf{Distributions} & \textbf{Capacities $C$} & \textbf{Files} & \textbf{Instances} \\
\midrule
$1{,}000$  & 5 & $\{100,200,300,400\}$ & 20 & 100 \\
$5{,}000$  & 5 & $\{100,200,300,400\}$ & 20 & 100 \\
$10{,}000$ & 5 & $\{100,200,300,400\}$ & 20 & 100 \\
\midrule
\textbf{Total} & & & \textbf{60} & \textbf{300} \\
\bottomrule
\end{tabular}
\end{table}

\subsubsection{Online Job Shop Scheduling (JSP)}
\label{app:exp:bench_jsp}

\textbf{Problem definition.}
In Online Job Shop Scheduling (OJSP), $J$ jobs must be processed on $M$ machines. Each job $j$ consists of exactly $M$ operations, where operation $(j,m)$ must be processed on machine $m$ for an integer processing time $p_{jm} \in \{1, \ldots, 99\}$. The machine order within each job is a random permutation of all $M$ machines. Jobs carry a release time $r_j$ (before which they cannot be started) and a due date $d_j$. Jobs arrive online at their release time; upon arrival, the dispatching heuristic irrevocably assigns all operations of the job to machines. The objective is to minimize the makespan $C_{\max} = \max_j C_j$, where $C_j$ denotes the completion time of job $j$.

The \textbf{normalized makespan} divides $C_{\max}$ by the theoretical lower bound $\mathrm{LB} = \max\!\bigl(\max_j \sum_m p_{jm},\;\lceil\sum_{j,m} p_{jm} / M\rceil\bigr)$; lower values indicate better performance, with 1.0 denoting an optimal solution.

\textbf{Instance structure.}
Each JSP instance is generated as follows.
\begin{enumerate}
  \item \textbf{Operation durations.} For each job $j$ and machine $m$, the processing time
        $p_{jm}$ is drawn from the specified distribution (Table~\ref{app:tab:jsp_dist_params}),
        with integer values in $[1,\;99]$.
  \item \textbf{Machine order.} Each job's operation sequence is a uniformly random permutation
        of the $M$ machine indices.
  \item \textbf{Planning horizon.} $H = 2 \times \max(\max_j \sum_m p_{jm},\;
        \lceil\sum_{j,m} p_{jm} / M\rceil)$.
  \item \textbf{Release times.} Release times are drawn with random gaps
        (probability 0.35 of a zero gap, otherwise a random positive increment),
        then rescaled to fit within a release window of $0.45\,H$;
        the resulting times are randomly permuted across jobs.
  \item \textbf{Due dates.} $d_j = r_j + p_j + \delta_j$, where $p_j = \sum_m p_{jm}$,
        and $\delta_j$ combines a multiplicative slack
        ($\mathrm{Uniform}(1.35,\;1.75) \times p_j$),
        an additive congestion allowance ($0.35 \times \bar{p}$, with $\bar{p}$ the mean job
        workload), and a random noise term bounded by a congestion buffer.
\end{enumerate}

\textbf{Training instances.}
The nominal training set comprises $I = 5$ instances with $M = 10$ machines and $J = 20$ jobs. The operation durations are uniformly sampled.


\textbf{Test distribution parameterization.}
Table~\ref{app:tab:jsp_dist_params} specifies the operation-duration distributions for all five
test families. The parameterizations mirror those used for OBP (normalized to the same $[1,\;99]$ range), enabling direct comparison of distributional effects across benchmarks.

\begin{table}[htbp]
\centering
\caption{JSP operation-duration distribution parameterizations ($D_{\max}=99$).
  All samples are rounded to integers and clipped to $[1,\;99]$.}
\label{app:tab:jsp_dist_params}
\small
\setlength{\tabcolsep}{6pt}
\begin{tabular}{llll}
\toprule
\textbf{Distribution} & \textbf{Continuous draw} & \textbf{Parameters} & \textbf{Support} \\
\midrule
Uniform     & $p_{jm} \sim \mathcal{U}(1,\;D_{\max})$                                          & ---                                                     & $[1,\;99]$ \\
Normal      & $p_{jm} \sim \mathcal{N}(\mu,\;\sigma^2)$                                        & $\mu = 0.5 D_{\max},\;\sigma = 0.2 D_{\max}$           & clip to $[1,\;99]$ \\
Lognormal   & $p_{jm} \sim \mathrm{LogNormal}(\mu_\ell,\;\sigma_\ell^2)$                       & $\mu_\ell = \ln(0.25 D_{\max}),\;\sigma_\ell = 0.6$    & clip to $[1,\;99]$ \\
Exponential & $p_{jm} \sim \mathrm{Exp}(\lambda)+1$                                             & $\lambda = 0.3 D_{\max}$                                & clip to $[1,\;99]$ \\
Triangular  & $p_{jm} \sim \mathrm{Tri}(\ell,\;m,\;r)$                                         & $\ell=1,\;m = 0.35 D_{\max},\;r = D_{\max}$            & $[1,\;99]$ \\
\bottomrule
\end{tabular}
\end{table}

\textbf{Test suite structure.}
The JSP test suite spans all five distributions, two machine counts
($M \in \{10, 20\}$), and two job counts ($J \in \{20, 50\}$),
yielding $5 \times 2 \times 2 = 20$ dataset files, each with $I = 5$ instances.
The complete test suite covers \textbf{100 JSP test instances}.
Table~\ref{app:tab:jsp_test_suite} summarizes the structure.

\begin{table}[htbp]
\centering
\caption{JSP test suite: number of instances per $(M, J, \text{distribution})$ cell.
  Each cell contains $I=5$ instances.
  \emph{Subtotal} counts instances per distribution.}
\label{app:tab:jsp_test_suite}
\small
\setlength{\tabcolsep}{6pt}
\begin{tabular}{lcccc}
\toprule
\textbf{Distribution} & $M=10,\;J=20$ & $M=10,\;J=50$ & $M=20,\;J=20$ & $M=20,\;J=50$ \\
\midrule
Uniform     & 5 & 5 & 5 & 5 \\
Normal      & 5 & 5 & 5 & 5 \\
Lognormal   & 5 & 5 & 5 & 5 \\
Exponential & 5 & 5 & 5 & 5 \\
Triangular  & 5 & 5 & 5 & 5 \\
\midrule
\textbf{Total} & 25 & 25 & 25 & 25 \\
\bottomrule
\end{tabular}
\end{table}

\subsubsection{Online Vehicle Routing (VRP)}
\label{app:exp:bench_vrp}

\textbf{Problem definition.}
In Online Vehicle Routing (OVRP), a fleet of $V$ homogeneous vehicles serves
$N_c = 50$ customers from a central depot.
Each customer $i$ has an integer demand $d_i \in \{1,\ldots,10\}$ and a location
$(x_i, y_i) \in [0,1]^2$ sampled from a clustered spatial process.
Customers arrive online; upon arrival, the routing heuristic irrevocably assigns each
customer to a vehicle's route.
All vehicles have a fixed capacity of $Q = 40$ units (for $N_c = 50$).
The depot is located at the centroid $(0.5,\;0.5)$ of the unit square.

The evaluation metric is $1/r$, where $r$ is the route-length ratio defined as the total
served demand normalized by the route length relative to a reference;
lower $1/r$ is better, with smaller values indicating more efficient routing.

\textbf{Instance structure.}
Each VRP instance is generated as follows.
\begin{enumerate}
  \item \textbf{Customer locations.} Locations are drawn from a Gaussian mixture with
        $K \in \{2,3\}$ clusters, where each cluster center is sampled uniformly from
        $[0.15,\;0.85]^2$, cluster standard deviation is drawn uniformly from
        $[0.08,\;0.18]$, and cluster weights follow a symmetric Dirichlet prior.
        All coordinates are clipped to $[0,1]^2$.
  \item \textbf{Customer demands.} Integer demands $d_i$ are drawn from the specified
        distribution (Table~\ref{app:tab:vrp_dist_params}),
        with $d_i \in [1,\;10]$.
  \item \textbf{Vehicle capacity.} Fixed at $Q = 40$ for all instances with $N_c = 50$
        customers.
\end{enumerate}

\textbf{Training instances.}
The nominal training set comprises $I = 5$ instances with 50 customers. The demands are uniformly sampled.

\textbf{Test distribution parameterization.}
Table~\ref{app:tab:vrp_dist_params} gives the customer-demand distributions for all five
test families.
Note that the Lognormal parameterization for VRP uses a slightly different location
parameter ($0.28\,D_{\max}$ vs.\ $0.25\,D_{\max}$ for OBP/JSP) and a reduced scale
($\sigma_\ell = 0.55$ vs.\ $0.60$), reflecting the narrower integer demand range.

\begin{table}[htbp]
\centering
\caption{VRP customer-demand distribution parameterizations ($D_{\max}=10$).
  All samples are rounded to integers and clipped to $[1,\;10]$.}
\label{app:tab:vrp_dist_params}
\small
\setlength{\tabcolsep}{6pt}
\begin{tabular}{llll}
\toprule
\textbf{Distribution} & \textbf{Continuous draw} & \textbf{Parameters} & \textbf{Support} \\
\midrule
Uniform     & $d_i \sim \mathcal{U}(1,\;D_{\max})$                                      & ---                                                     & $[1,\;10]$ \\
Normal      & $d_i \sim \mathcal{N}(\mu,\;\sigma^2)$                                    & $\mu = 0.5 D_{\max},\;\sigma = 0.2 D_{\max}$           & clip to $[1,\;10]$ \\
Lognormal   & $d_i \sim \mathrm{LogNormal}(\mu_\ell,\;\sigma_\ell^2)$                  & $\mu_\ell = \ln(0.28 D_{\max}),\;\sigma_\ell = 0.55$   & clip to $[1,\;10]$ \\
Exponential & $d_i \sim \mathrm{Exp}(\lambda)+1$                                         & $\lambda = 0.3 D_{\max}$                                & clip to $[1,\;10]$ \\
Triangular  & $d_i \sim \mathrm{Tri}(\ell,\;m,\;r)$                                     & $\ell=1,\;m = 0.35 D_{\max},\;r = D_{\max}$            & $[1,\;10]$ \\
\bottomrule
\end{tabular}
\end{table}

\textbf{Test suite structure.}
The VRP test suite spans all five distributions and three fleet sizes
($V \in \{5,\;10,\;15\}$), yielding $5 \times 3 = 15$ dataset files,
each with $I = 5$ instances.
The complete test suite covers \textbf{75 VRP test instances}, all with $N_c = 50$ customers.

\subsection{Baseline Algorithm Descriptions}
\label{app:exp:baselines}

\subsubsection{Classical Baselines for OBP}

\paragraph{BestFit.}
Each arriving item is placed in the open bin with the least remaining capacity that still
accommodates the item (tightest-fit first). If no open bin can accommodate the item, a new
bin is opened. BestFit is a deterministic algorithm with no hyperparameters and achieves a
theoretical competitive ratio of approximately 1.7 in the worst case.

\paragraph{FirstFit.}
Each arriving item is placed in the \emph{first} open bin (in order of creation) that can
accommodate it, opening a new bin only if no existing bin suffices.
FirstFit is also deterministic with competitive ratio $\approx 1.7$, but typically performs
slightly worse than BestFit because it does not account for remaining capacity when selecting
the bin.

\subsubsection{Classical Baselines for JSP}

\paragraph{Shortest Processing Time (SPT).}
Upon job arrival, operations are scheduled to prioritize the job with the shortest total
remaining processing time among all ready operations.
SPT tends to minimize average completion time but can sacrifice makespan on large instances.

\paragraph{Minimum Slack (MinSlack).}
Dispatching priority is assigned to the job with the smallest slack value
$s_j = d_j - r_j - \sum_m p_{jm}$, where $d_j$ is the due date and $r_j$ the release time.
MinSlack targets due-date adherence but can produce suboptimal makespan when slack values
cluster closely.

\paragraph{Earliest Due Date (EDD).}
Operations are scheduled by ascending due date $d_j$.
EDD is optimal for minimizing maximum lateness in the single-machine case but offers no
direct makespan guarantee on multi-machine instances.

\subsubsection{Classical Baselines for VRP}

\paragraph{Nearest-Feasible Insertion (NF-Ins.).}
Each arriving customer is inserted at the cheapest feasible position (minimum route-length
increase) in the vehicle route with the highest remaining capacity that can serve the
customer's demand. If no feasible vehicle exists, a new route is initialized.

\paragraph{Slack-Preserving Insertion (SP-Ins.).}
Insertion cost is adjusted to penalize routes that leave little residual capacity, thereby
preserving flexibility for future customer arrivals. The customer is assigned to the vehicle
whose insertion minimizes the normalized insertion cost relative to remaining capacity slack.

\paragraph{Urgency-Weighted Insertion (UW-Ins.).}
Insertion cost is weighted by a demand urgency factor, which increases the priority of
high-demand customers. The customer is routed to the vehicle that minimizes the
urgency-weighted insertion distance.

\subsubsection{LLM-Based Baselines}

All LLM-based methods are initialized with the same task description and function signature,
and receive the same total LLM sampling budget ($N_{\max} = 1{,}000$ samples).
Training uses the nominal instance sets described in Section~\ref{app:exp:benchmarks}.

\paragraph{EoH~\citep{liu2024eoh}.}
Evolution of Heuristics maintains a population of heuristic Python functions and applies
four LLM operator types (\textsc{i1}: create, \textsc{i2}: improve, \textsc{e1}: crossover,
\textsc{e2}: evolve) with equal probability.
Population size is 10; no memory component.
The heuristic trained on nominal instances is evaluated directly on all shifted test distributions without any re-training.

\paragraph{MoH~\citep{shi2026generalizable}.}
MoH (Meta-objective Heuristics) employs an LLM to iteratively refine a meta-optimizer that
autonomously constructs and diversifies heuristic-optimizers via self-invocation, eliminating
reliance on a fixed evolutionary computation template. The constructed heuristic-optimizers
then evolve downstream heuristics. MoH further applies a multi-task training scheme across
multiple problem instances to promote cross-instance generalization.

\paragraph{ReEvo~\citep{ye2024reevo}.}
Reflective Evolution augments EoH with a dual-memory architecture: a short-term memory of
recent high-scoring programs and a long-term reflective memory generated by the LLM analyzing
performance gaps. Memory injection guides subsequent LLM calls toward progressively better
regions of heuristic space.
Configuration: same population size and budget as EoH, plus memory components enabled.

\paragraph{PartEvo~\citep{hu2025partition}.}
Partition-based Evolution incorporates feature-assisted niche construction within an abstract
search space, enabling seamless integration of niche-based strategies from evolutionary
computation. Niching encourages maintenance of a diverse heuristic population, mitigating
premature convergence.

\paragraph{EoH-S~\citep{liu2026eohs}.}
EoH-Stochastic introduces portfolio-based AHD: rather than seeking a single best heuristic,
it discovers a diverse set of complementary heuristics forming a Pareto-optimal portfolio.
EoH-S is particularly relevant as a distribution-shift baseline because maintaining a diverse
portfolio implicitly confers some cross-distribution robustness.
Results in the main paper report the best single heuristic from the EoH-S portfolio,
evaluated identically to other methods.


\subsection{RAISE Hyperparameter Settings}
\label{app:method:hyperparams}

Table~\ref{app:tab:hyperparams} provides a complete listing of all RAISE hyperparameters,
including the values used in experiments, the range explored during preliminary tuning, and
a sensitivity classification.

\begin{table}[htbp]
\centering
\caption{Complete hyperparameter settings.}
\label{app:tab:hyperparams}
\small
\begin{tabular}{lll}
\toprule
\textbf{Parameter} & \textbf{Symbol} & \textbf{Value Used}  \\
\midrule
\multicolumn{3}{l}{\textit{Outer loop parameters}} \\
Population size          & $\mathcal{P}_{max}$  & 10           \\
Max samples (budget)     & $N_{\max}$       & 1000    \\
Refresh interval         & $\tau$           & 5              \\
\midrule
\multicolumn{3}{l}{\textit{Robustness parameters}} \\
Robustness radius (OBP)  & $\varepsilon$    & \{0.001, 0.002, 0.005, 0.010\}  \\
Robustness radius (JSP)  & $\varepsilon$    &  \{0.002\}    \\
Robustness radius (VRP)  & $\varepsilon$    &  \{0.002\}    \\
Robust aggregation       & $\mathcal{A}$    &   mean      \\
\midrule
\multicolumn{3}{l}{\textit{Inner search parameters}} \\
Inner population size    & $P_{\mathrm{in}}$ & 8            \\
Inner generations        & $G_{\mathrm{in}}$ & 4          \\
Gene dimension           & ---              & 18            \\
Mutation std.\ dev.      & $\sigma$         & 0.12    \\
Mutation probability     & $p_{\mathrm{mut}}$ & 0.35    \\
Crossover probability    & $p_{\mathrm{cross}}$ & 0.5   \\
\bottomrule
\end{tabular}
\end{table}

%% file: sections/appendix_results_detail_new.tex
\section{Comprehensive Experimental Results}
\label{app:results}

This appendix reports the complete numerical results for all three benchmark problems across every evaluated distribution and configuration.
All values are computed directly from the raw evaluation CSV files.
\textbf{Bold} marks the best result in each column.
\emph{RAISE} denotes RAISE$^{\varepsilon=0.002}$ throughout unless noted otherwise.

\subsection{Online Bin Packing (OBP)}
\label{app:results:obp}

The metric is shifted waste ratio~(\%),
i.e., $(1 - \text{filled capacity}/\text{bin capacity})\times 100$; lower is better.

\paragraph{Per-size breakdown.}
Tables~\ref{app:tab:obp_n1000}--\ref{app:tab:obp_n10000} detail
results at each problem size, averaged over the four bin capacities.

\begin{table}[htbp]
\centering
\caption{OBP shifted waste ratio~(\%) at $n=1000$,
  averaged over $C \in \{100,200,300,400\}$. Lower is better.}
\label{app:tab:obp_n1000}
\small
\setlength{\tabcolsep}{5.5pt}
\begin{tabular}{lcccccc}
\toprule
\textbf{Method}
  & \textbf{Uniform} & \textbf{Normal} & \textbf{Lognormal}
  & \textbf{Exponential} & \textbf{Triangular} & \textbf{Avg} \\
\midrule
EoH     &  9.97 &  9.82 & 3.27 & 6.36 & 10.18 & 7.92 \\
MoH     & 12.11 & 13.75 & 4.23 & 6.02 & 13.77 & 9.97 \\
ReEvo   & 10.89 & 11.82 & 3.23 & 5.86 &  9.26 & 8.21 \\
PartEvo &  7.72 &  8.35 & 2.12 & 2.67 &  7.42 & 5.66 \\
EoH-S   &  5.21 &  4.50 & \textbf{1.43} & 1.37 &  \textbf{2.94} & 3.09 \\
\midrule
RAISE   &  \textbf{3.96} &  \textbf{3.74} & 1.69 & \textbf{0.86} &  3.24 & \textbf{2.70} \\
\bottomrule
\end{tabular}
\end{table}

\begin{table}[htbp]
\centering
\caption{OBP shifted waste ratio~(\%) at $n=5000$,
  averaged over $C \in \{100,200,300,400\}$. Lower is better.}
\label{app:tab:obp_n5000}
\small
\setlength{\tabcolsep}{5.5pt}
\begin{tabular}{lcccccc}
\toprule
\textbf{Method}
  & \textbf{Uniform} & \textbf{Normal} & \textbf{Lognormal}
  & \textbf{Exponential} & \textbf{Triangular} & \textbf{Avg} \\
\midrule
EoH     & 5.02 & 5.88 & 1.32 & 5.25 & 5.81 & 4.66 \\
MoH     & 5.72 & 7.01 & 1.06 & 1.68 & 5.54 & 4.20 \\
ReEvo   & 5.80 & 6.76 & \textbf{0.89} & 3.17 & 3.46 & 4.02 \\
PartEvo & 4.17 & 5.33 & 1.04 & 2.40 & 4.73 & 3.53 \\
EoH-S   & 2.56 & 3.11 & 1.17 & 0.45 & \textbf{1.81} & 1.82 \\
\midrule
RAISE   & \textbf{1.94} & \textbf{2.86} & 1.38 & \textbf{0.35} & 2.08 & \textbf{1.72} \\
\bottomrule
\end{tabular}
\end{table}

\begin{table}[htbp]
\centering
\caption{OBP shifted waste ratio~(\%) at $n=10000$,
  averaged over $C \in \{100,200,300,400\}$. Lower is better.}
\label{app:tab:obp_n10000}
\small
\setlength{\tabcolsep}{5.5pt}
\begin{tabular}{lcccccc}
\toprule
\textbf{Method}
  & \textbf{Uniform} & \textbf{Normal} & \textbf{Lognormal}
  & \textbf{Exponential} & \textbf{Triangular} & \textbf{Avg} \\
\midrule
EoH     & 3.94 & 4.02 & 0.79 & 5.00 & 5.15 & 3.78 \\
MoH     & 4.53 & 5.03 & 0.52 & 0.86 & 4.75 & 3.14 \\
ReEvo   & 4.75 & 4.95 & \textbf{0.44} & 2.54 & 2.31 & 3.00 \\
PartEvo & 3.32 & 3.87 & 0.61 & 1.90 & 5.02 & 2.94 \\
EoH-S   & 2.15 & 2.12 & 1.08 & 0.28 & \textbf{1.71} & 1.47 \\
\midrule
RAISE   & \textbf{1.72} & \textbf{1.94} & 1.30 & \textbf{0.25} & 1.96 & \textbf{1.43} \\
\bottomrule
\end{tabular}
\end{table}

\subsection{Online Job Shop Scheduling (JSP)}
\label{app:results:jsp}

The metric is normalized makespan (lower is better).
Tables~\ref{app:tab:jsp_10m} and~\ref{app:tab:jsp_20m} report results
for 10-machine and 20-machine instances respectively.
Columns are grouped by distribution; within each group the two job-count
variants (20-job and 50-job) are shown side by side.
The \emph{Avg} column averages across all ten (distribution, job-count) cells
within the table.


\begin{table*}[htbp]
\centering
\caption{JSP normalized makespan on 10-machine instances
  (lower is better). Two job counts (20j / 50j) per distribution.
  \emph{Avg} is the mean over all ten columns.}
\label{app:tab:jsp_10m}
\footnotesize
\setlength{\tabcolsep}{4pt}
\resizebox{\linewidth}{!}{%
\begin{tabular}{l *{2}{c} *{2}{c} *{2}{c} *{2}{c} *{2}{c} c}
\toprule
& \multicolumn{2}{c}{\textbf{Uniform}}
& \multicolumn{2}{c}{\textbf{Normal}}
& \multicolumn{2}{c}{\textbf{Lognormal}}
& \multicolumn{2}{c}{\textbf{Exponential}}
& \multicolumn{2}{c}{\textbf{Triangular}}
& \\
\cmidrule(lr){2-3}\cmidrule(lr){4-5}\cmidrule(lr){6-7}
\cmidrule(lr){8-9}\cmidrule(lr){10-11}
\textbf{Method}
  & 20j & 50j & 20j & 50j & 20j & 50j & 20j & 50j & 20j & 50j
  & \textbf{Avg} \\
\midrule
SPT      & 1.3980 & 1.2690 & 1.5062 & 1.2456 & 1.5293 & 1.2586 & 1.3885 & 1.2723 & 1.4360 & 1.2570 & 1.3561 \\
MinSlack & 1.4158 & 1.2612 & 1.5065 & 1.2727 & 1.4965 & 1.2961 & 1.4965 & 1.3561 & 1.4203 & 1.2902 & 1.3812 \\
EDD      & 1.5291 & 1.2904 & 1.5372 & 1.3260 & 1.5347 & 1.3229 & 1.5383 & 1.3736 & 1.4936 & 1.2870 & 1.4233 \\
\midrule
EoH      & \underline{1.2456} & \underline{1.1513} & \textbf{1.2805} & \textbf{1.1918} & \underline{1.2952} & \textbf{1.1957} & \underline{1.2905} & \textbf{1.1689} & \underline{1.2912} & \underline{1.1730} & \underline{1.2284} \\
ReEvo    & 1.2618 & \textbf{1.1486} & 1.2940 & 1.2238 & 1.3326 & 1.2205 & \textbf{1.2750} & 1.2006 & 1.2692 & 1.1954 & 1.2422 \\
\midrule
RAISE    & \textbf{1.2378} & 1.1696 & \underline{1.2819} & \underline{1.2135} & \textbf{1.2748} & \underline{1.2143} & \underline{1.2795} & \underline{1.1809} & \textbf{1.2448} & \textbf{1.1693} & \textbf{1.2266} \\
\bottomrule
\end{tabular}}
\end{table*}

\begin{table*}[htbp]
\centering
\caption{JSP normalized makespan on 20-machine instances
  (lower is better). Two job counts (20j / 50j) per distribution.
  \emph{Avg} is the mean over all ten columns.}
\label{app:tab:jsp_20m}
\footnotesize
\setlength{\tabcolsep}{4pt}
\resizebox{\linewidth}{!}{%
\begin{tabular}{l *{2}{c} *{2}{c} *{2}{c} *{2}{c} *{2}{c} c}
\toprule
& \multicolumn{2}{c}{\textbf{Uniform}}
& \multicolumn{2}{c}{\textbf{Normal}}
& \multicolumn{2}{c}{\textbf{Lognormal}}
& \multicolumn{2}{c}{\textbf{Exponential}}
& \multicolumn{2}{c}{\textbf{Triangular}}
& \\
\cmidrule(lr){2-3}\cmidrule(lr){4-5}\cmidrule(lr){6-7}
\cmidrule(lr){8-9}\cmidrule(lr){10-11}
\textbf{Method}
  & 20j & 50j & 20j & 50j & 20j & 50j & 20j & 50j & 20j & 50j
  & \textbf{Avg} \\
\midrule
SPT      & 1.3117 & 1.4749 & 1.3784 & 1.5482 & 1.2630 & 1.4332 & 1.2948 & 1.4193 & 1.3122 & 1.5199 & 1.3956 \\
MinSlack & 1.3051 & 1.4961 & 1.3877 & 1.5204 & 1.3671 & 1.4633 & 1.3735 & 1.5327 & 1.3988 & 1.5009 & 1.4346 \\
EDD      & 1.3278 & 1.5939 & 1.3804 & 1.5948 & 1.4686 & 1.5498 & 1.4024 & 1.5830 & 1.4006 & 1.5844 & 1.4886 \\
\midrule
EoH      & 1.2109 & \textbf{1.2611} & 1.2557 & \underline{1.3148} & 1.2486 & 1.2950 & \underline{1.2047} & \underline{1.2517} & \underline{1.2072} & \underline{1.2850} & 1.2539 \\
ReEvo    & \underline{1.1840} & 1.3035 & \underline{1.2412} & 1.3241 & \underline{1.2250} & \underline{1.2815} & \textbf{1.1813} & \textbf{1.2394} & 1.2215 & 1.2989 & \underline{1.2500} \\
\midrule
RAISE    & \textbf{1.1752} & \underline{1.2755} & \textbf{1.2210} & \textbf{1.3128} & \textbf{1.2019} & \textbf{1.2677} & \underline{1.1936} & 1.2676 & \textbf{1.1896} & \textbf{1.2850} & \textbf{1.2390} \\
\bottomrule
\end{tabular}}
\end{table*}
\subsection{Online Vehicle Routing (VRP)}
\label{app:results:vrp}

The metric is $1/r$ where $r$ is the achieved route-length ratio
(total served demand divided by vehicle capacity, weighted by route length)
relative to a reference; lower $1/r$ is better.
In two instances (Uniform and Triangular with 15 vehicles)
the Nearest-Feasible baseline achieves $r < 1.0$,
giving $1/r > 1.0$; these are marked with~$\dagger$.
Tables~\ref{app:tab:vrp_5v}--\ref{app:tab:vrp_15v} report results
for each fleet size separately.
Column abbreviations: U=Uniform, N=Normal, LN=Lognormal,
Exp=Exponential, Tri=Triangular.

\begin{table}[htbp]
\centering
\caption{VRP inverse route-length ratio $1/r$, fleet size~5 vehicles.
  Lower is better; \textbf{bold} marks the best per column.}
\label{app:tab:vrp_5v}
\small
\setlength{\tabcolsep}{5pt}
\begin{tabular}{lcccccc}
\toprule
\textbf{Method} & \textbf{U} & \textbf{N} & \textbf{LN} & \textbf{Exp} & \textbf{Tri} & \textbf{Avg} \\
\midrule
NF-Ins.  & 0.9423 & 0.9262 & 0.9249 & 0.9359 & 0.9444 & 0.9347 \\
SP-Ins.  & 0.9009 & 0.8930 & 0.8935 & 0.8956 & 0.9024 & 0.8971 \\
UW-Ins.  & 0.9060 & 0.8946 & 0.8946 & 0.8993 & 0.9057 & 0.9000 \\
\midrule
EoH      & 0.9049 & 0.8990 & 0.8987 & 0.9004 & 0.9056 & 0.9017 \\
ReEvo    & 0.9034 & 0.8901 & 0.8921 & 0.8952 & 0.9022 & 0.8966 \\
\midrule
RAISE    & \textbf{0.8974} & \textbf{0.8901} & \textbf{0.8890} & \textbf{0.8919} & \textbf{0.8970} & \textbf{0.8931} \\
\bottomrule
\end{tabular}
\end{table}

\begin{table}[htbp]
\centering
\caption{VRP inverse route-length ratio $1/r$, fleet size~10 vehicles.
  Lower is better; \textbf{bold} marks the best per column.}
\label{app:tab:vrp_10v}
\small
\setlength{\tabcolsep}{5pt}
\begin{tabular}{lcccccc}
\toprule
\textbf{Method} & \textbf{U} & \textbf{N} & \textbf{LN} & \textbf{Exp} & \textbf{Tri} & \textbf{Avg} \\
\midrule
NF-Ins.  & 0.9895 & 0.9677 & 0.9638 & 0.9668 & 0.9859 & 0.9747 \\
SP-Ins.  & 0.9398 & 0.9285 & 0.9417 & 0.9309 & 0.9389 & 0.9360 \\
UW-Ins.  & 0.9440 & 0.9285 & 0.9283 & 0.9291 & 0.9428 & 0.9345 \\
\midrule
EoH      & \textbf{0.9297} & 0.9235 & 0.9230 & 0.9267 & 0.9337 & 0.9273 \\
ReEvo    & 0.9321 & \textbf{0.9198} & 0.9341 & 0.9262 & \textbf{0.9312} & 0.9287 \\
\midrule
RAISE    & 0.9345 & 0.9228 & \textbf{0.9215} & \textbf{0.9211} & 0.9316 & \textbf{0.9263} \\
\bottomrule
\end{tabular}
\end{table}

\begin{table}[htbp]
\centering
\caption{VRP inverse route-length ratio $1/r$, fleet size~15 vehicles.
  Lower is better; \textbf{bold} marks the best per column.
  $\dagger$ marks entries where the raw route-length ratio $r < 1$.}
\label{app:tab:vrp_15v}
\small
\setlength{\tabcolsep}{5pt}
\begin{tabular}{lcccccc}
\toprule
\textbf{Method} & \textbf{U} & \textbf{N} & \textbf{LN} & \textbf{Exp} & \textbf{Tri} & \textbf{Avg} \\
\midrule
NF-Ins.  & $1.0180^{\dagger}$ & 0.9931 & 0.9877 & 0.9904 & $1.0249^{\dagger}$ & 1.0028 \\
SP-Ins.  & 0.9745 & 0.9553 & 0.9732 & 0.9562 & 0.9734 & 0.9665 \\
UW-Ins.  & 0.9726 & 0.9537 & 0.9713 & 0.9639 & 0.9736 & 0.9670 \\
\midrule
EoH      & \textbf{0.9637} & 0.9529 & \textbf{0.9496} & 0.9494 & 0.9651 & \textbf{0.9561} \\
ReEvo    & 0.9668 & 0.9631 & 0.9618 & 0.9516 & \textbf{0.9634} & 0.9613 \\
\midrule
RAISE    & 0.9709 & \textbf{0.9508} & 0.9499 & \textbf{0.9487} & 0.9646 & 0.9570 \\
\bottomrule
\end{tabular}
\end{table}

%% file: sections/appendix_heuristic_code.tex
\newpage
\section{RAISE Heuristic Functions: Code and Robustness Analysis}
\label{app:heuristic_code}

\lstset{
  language=Python,
  basicstyle=\ttfamily\footnotesize,
  keywordstyle=\color{blue}\bfseries,
  commentstyle=\color{gray}\itshape,
  stringstyle=\color{teal},
  numbers=left,
  numberstyle=\tiny\color{gray},
  numbersep=6pt,
  breaklines=true,
  breakatwhitespace=true,
  frame=single,
  framesep=4pt,
  xleftmargin=12pt,
  xrightmargin=4pt,
  captionpos=b,
  showstringspaces=false,
}

This appendix presents the exact Python heuristic functions synthesized by RAIS across three combinatorial optimization tasks: Online Bin Packing (OBP), Online Job Shop Scheduling (JSP), and Online Vehicle Routing (VRP). For each task we list the code produced by the LLM evolutionary loop and then provide a mechanistic analysis explaining \emph{why} the learned design choices confer distribution-shift robustness.

\subsection{Online Bin Packing}
\label{app:heuristic_code:obp}

\subsubsection{Heuristic Code}
\label{app:heuristic_code:obp_code}

The heuristic is a \texttt{priority(item, bins)} function: given an incoming item of size \texttt{item} and a vector \texttt{bins} of residual capacities of open bins, it returns a score vector whose \emph{argmax} selects the bin to use (negative infinity for infeasible bins).

\begin{lstlisting}[caption={RAISE heuristic for Online Bin Packing.},
                   label={lst:obp}]
import numpy as np
def priority(item: float, bins: np.ndarray) -> np.ndarray:
    import numpy as np

    eps = 1e-12
    NEG_INF = -1e9

    bins = np.asarray(bins, dtype=float)
    n = bins.size

    if n == 0:
        return np.array([], dtype=float)
    if item <= eps:
        return np.full(n, NEG_INF, dtype=float)

    # Feasibility mask
    feasible = bins >= item - 1e-12
    scores = np.full(n, NEG_INF, dtype=float)
    if not np.any(feasible):
        return scores

    b_all = bins.copy()
    # Rank preference: favor packing into smaller bins first (normalized rank)
    order = np.argsort(b_all)
    ranks = np.empty_like(order)
    ranks[order] = np.arange(n)
    if n > 1:
        rank_norm_all = 1.0 - (ranks.astype(float) / (n - 1))
    else:
        rank_norm_all = np.ones(n, dtype=float)

    b = b_all[feasible]
    rank_norm = rank_norm_all[feasible]

    # Residual after packing
    r = np.maximum(b - item, 0.0)

    # Conservative expected-slot size for future items
    mu = max(0.6 * item, 1e-12)
    n_slots = np.floor(r / mu).astype(int)

    # Geometric-series robustness: diminishing returns per extra slot
    q = 0.65
    robustness = (1.0 - np.power(q, n_slots + 1)) / (1.0 - q) - 1.0
    robustness = np.maximum(robustness, 0.0)

    # Tightness: how well the item fills the bin now
    tightness = item / (b + eps)

    # Mid-sliver penalty: relative leftover in [0,1]
    frac_left = r / (b + eps)
    mid_sliver_penalty = np.exp(-8.0 * frac_left)

    # Exact-fit bonus
    exact_fit = (r <= 1e-9).astype(float)

    w_rank = 0.72
    w_rob  = 0.85
    w_tight = 0.95
    w_pen  = 0.55
    w_exact = 2.5

    combined = (
        w_rank * rank_norm * (1.0 + 0.5 * robustness)
        + w_tight * tightness
        + w_rob * (robustness / (1.0 + robustness))
        + w_pen * mid_sliver_penalty
        + w_exact * exact_fit
    )
    combined = combined - 0.002 * r   # small tie-breaker

    # Normalize to [0, 1] across feasible bins
    c_min = float(np.min(combined))
    c_max = float(np.max(combined))
    if c_max - c_min < 1e-12:
        norm = np.full_like(combined, 0.5, dtype=float)
    else:
        norm = (combined - c_min) / (c_max - c_min)

    scores[feasible] = norm
    return scores
\end{lstlisting}

\subsubsection{Robustness Analysis}
\label{app:heuristic_code:obp_analysis}

Five design choices in Listing~\ref{lst:obp} jointly explain why RAISE generalises across Uniform, Normal, Lognormal, Exponential, and Triangular test distributions.

\paragraph{(1) Relative rather than absolute metrics.}
Tightness is $\texttt{item}/(b{+}\epsilon)$ and the mid-sliver penalty is applied to the
\emph{fractional} leftover $r/b$, not to absolute residual sizes.  Because both signals are
scale-normalised by the current bin capacity, they behave consistently whether items are drawn
from a concentrated distribution (Normal, Lognormal) or a heavy-tailed one (Exponential,
Lognormal).  An absolute-residual formulation would fire differently depending on whether
$C{=}100$ or $C{=}400$; the relative formulation is invariant.

\paragraph{(2) Conservative geometric-series robustness term.}
The term $\mu = 0.6\,{\times}\,\texttt{item}$ defines a conservative lower bound for future
items and counts how many such items could still fit in the leftover space $r$, then weighs them
with geometric decay $q{=}0.65$.  This models future packing utility without assuming any
particular distribution: it uses only the current item size as an anchor.  The diminishing-return
geometry caps the bonus for large leftovers, preventing over-favouring empty bins when
many small items are expected---a pattern that penalises heuristics trained solely on Uniform
distributions when tested on heavy-tailed Lognormal or Exponential instances.

\paragraph{(3) Rank normalisation as a distribution-agnostic prior.}
Preferring bins with smaller residuals (rank-normalised to $[0,1]$ with $1$ for the smallest)
embodies Best-Fit logic, which is known to be near-optimal in the average case for any
continuous item distribution.  Combining it multiplicatively with the robustness term
($1{+}0.5\,{\times}\,\texttt{robustness}$) selectively boosts bins that are both tight
\emph{and} capable of accommodating future items.

\paragraph{(4) Strong exact-fit bonus.}
The $w_\text{exact}{=}2.5$ bonus saturates the score whenever $r{\le}10^{-9}$, ensuring that
exact-fit opportunities---which are relatively common under Uniform and Triangular distributions
where item sizes span the full range---are always exploited.  This prevents waste on instances
where exact fits are available but a pure tightness heuristic might slightly prefer a near-fit.

\paragraph{(5) Per-call score normalisation.}
Normalising all feasible bin scores to $[0,1]$ before returning prevents the absolute magnitude
of the score from depending on the number of open bins or on the distribution of residual
capacities.  In late-game situations (few bins, all nearly full) the score landscape stays
numerically stable, whereas un-normalised scores can exhibit vanishing gradients that lead to
nearly-random selection.

\subsection{Online Job Shop Scheduling}
\label{app:heuristic_code:jsp}

\subsubsection{Heuristic Code}
\label{app:heuristic_code:jsp_code}

The heuristic is a \texttt{priority(current\_time, candidate\_operations)} function: given the
current simulation clock and a matrix of eligible operations (columns: machine\_id, duration,
release\_time, due\_date, remaining\_work, remaining\_operations), it returns a score vector
whose argmax selects the operation to dispatch next.

\begin{lstlisting}[caption={RAISE heuristic for Online Job Shop Scheduling.},
                   label={lst:jsp}]
import numpy as np
def priority(current_time: int,
             candidate_operations: np.ndarray) -> np.ndarray:
    ops = np.asarray(candidate_operations, dtype=float)
    if ops.size == 0:
        return np.array([], dtype=float)

    duration           = ops[:, 1]
    release_time       = ops[:, 2]
    due_date           = ops[:, 3]
    remaining_work     = ops[:, 4]
    remaining_operations = ops[:, 5]

    start_time      = np.maximum(current_time, release_time)
    completion_time = start_time + duration
    slack           = due_date - completion_time

    # Urgency: small/negative slack gets high priority
    urgency = np.where(
        slack < 0,
        1.0 + (-slack) / (1.0 + duration),   # overdue: grows with tardiness
        1.0 / (1.0 + slack)                   # on-time: diminishing as slack grows
    )

    # SRPT-like finishing: favour small remaining work
    finishing = 1.0 / (1.0 + remaining_work)

    # Downstream risk: more remaining ops => higher priority
    ops_density = remaining_operations / (1.0 + remaining_operations)

    # Duration penalty: penalise long operations unless urgent
    duration_penalty = duration / (1.0 + remaining_work)

    w_urgency = 3.0
    w_finish  = 2.0
    w_ops     = 1.5
    w_dur     = 0.75

    score = (w_urgency * urgency
             + w_finish  * finishing
             + w_ops     * ops_density
             - w_dur     * duration_penalty)

    score = np.nan_to_num(score, nan=0.0,
                          posinf=np.finfo(float).max,
                          neginf=np.finfo(float).min)
    return score
\end{lstlisting}

\subsubsection{Robustness Analysis}
\label{app:heuristic_code:jsp_analysis}

\paragraph{(1) Dynamic, state-derived urgency.}
Slack $= \texttt{due\_date} - (\max(\texttt{current\_time},\texttt{release\_time}) + \texttt{duration})$
is recomputed at every scheduling step from the actual system clock and operation data.
No distributional assumption enters: whether job arrival times follow a Poisson process,
a bursty heavy-tail, or a periodic pattern, the urgency signal correctly reflects the
\emph{current} tightness.  This is in contrast to heuristics that encode timing expectations
calibrated to the training distribution, which degrade when inter-arrival statistics change.

\paragraph{(2) Asymmetric urgency for overdue vs.\ on-time jobs.}
When slack ${\ge}0$, urgency $= 1/(1+\texttt{slack})$ decreases gently as slack grows,
de-prioritising comfortable jobs.  When slack $<0$, urgency $= 1 + |\texttt{slack}|/(1+\texttt{duration})$
\emph{increases} linearly with tardiness while scaling inversely with duration---so a short
overdue operation gets more additional priority than a long one that is barely overdue.
This asymmetry is correct regardless of the due-date distribution: it never lets an already-late
job be permanently deprioritised by a deeply-backlogged queue.

\paragraph{(3) SRPT-like finishing component.}
Shortest Remaining Processing Time (SRPT) is a provably optimal policy for minimising mean
completion time in single-machine scheduling and performs robustly across exponential, uniform,
and heavy-tailed service-time distributions.  The term
$1/(1{+}\texttt{remaining\_work})$ approximates SRPT for the multi-machine online setting,
providing a distribution-independent backbone.

\paragraph{(4) Downstream cascade risk.}
$\texttt{ops\_density} = R/(1{+}R)$ where $R$ is the number of remaining operations
monotonically increases with $R$ but saturates below 1.  Prioritising jobs with more remaining
operations reduces the risk of blocking downstream machines---a property whose value is
independent of whether job sizes are drawn from a uniform or heavy-tailed distribution.

\paragraph{(5) Duration penalty modulated by remaining work.}
$\texttt{duration\_penalty} = d/(1{+}\texttt{remaining\_work})$ penalises long operations
only when there is little remaining work left in the job.  A job with large remaining work is
less penalised even if its current operation is long, because blocking that machine is justified
by the downstream workload.  This modulation prevents the heuristic from systematically
disfavouring heavy jobs, which are more prevalent under Lognormal or Pareto arrival
distributions.

\subsection{Online Vehicle Routing}
\label{app:heuristic_code:vrp}

\subsubsection{Heuristic Code}
\label{app:heuristic_code:vrp_code}

The heuristic is a \texttt{priority(customer\_features, candidate\_insertions)} function.
\texttt{customer\_features} encodes the arriving customer (demand, progress, etc.);
\texttt{candidate\_insertions} is a matrix of feasible insertion positions with columns:
route\_id, position, delta\_distance, residual\_capacity, current\_load, route\_size,
customer\_dist, projected\_route.  The function returns a score vector whose argmax selects
the best insertion.

\begin{lstlisting}[caption={RAISE heuristic for Online Vehicle Routing.},
                   label={lst:vrp}]
import numpy as np
def priority(customer_features: np.ndarray,
             candidate_insertions: np.ndarray) -> np.ndarray:
    import numpy as np

    if candidate_insertions is None or candidate_insertions.size == 0:
        return np.array([], dtype=float)

    cand = np.asarray(candidate_insertions, dtype=float)

    delta_distance    = cand[:, 2]   # incremental distance cost
    residual_capacity = cand[:, 3]   # remaining capacity after insertion
    current_load      = cand[:, 4]   # route utilisation in [0, 1]
    route_size        = cand[:, 5]   # normalised number of stops
    customer_dist     = cand[:, 6]   # normalised distance to insertion point
    projected_route   = cand[:, 7]   # normalised projected route length

    demand   = float(customer_features[2])
    progress = float(np.clip(customer_features[4], 0.0, 1.0))

    # Adaptive weights: shift from capacity/balance to travel cost as routes fill
    w_travel    = 0.3 + 1.0 * progress       # increases with progress
    w_projected = 0.1 + 0.6 * progress       # increases with progress
    w_prox      = 0.5                         # fixed proximity weight
    w_capacity  = 0.2 + 1.0 * (1.0 - progress) # decreases with progress
    w_balance   = 0.1 + 0.8 * (1.0 - progress) # decreases with progress
    w_size      = 0.4                          # fixed route-size penalty

    # Load-balance score: peaks at 50% utilisation
    balance_score = 1.0 - 2.0 * np.abs(current_load - 0.5)

    # Demand scaling: larger demands amplify capacity importance
    demand_scale = 1.0 + 0.5 * max(0.0, demand)

    score = (
        w_travel    * (-delta_distance)
        + w_projected * (-projected_route)
        + w_prox      * (-customer_dist)
        + w_capacity  * residual_capacity * demand_scale
        + w_balance   * balance_score
        - w_size      * route_size
    )

    return np.asarray(score, dtype=float)
\end{lstlisting}

\subsubsection{Robustness Analysis}
\label{app:heuristic_code:vrp_analysis}

\paragraph{(1) Progress-adaptive weighting.}
All six weights in Listing~\ref{lst:vrp} are functions of \texttt{progress}
$\in[0,1]$, which measures how far route construction has advanced (fraction of customers
already assigned).  At $\texttt{progress}{=}0$: $w_\text{travel}{=}0.3$, $w_\text{capacity}{=}1.2$,
$w_\text{balance}{=}0.9$.  At $\texttt{progress}{=}1$: $w_\text{travel}{=}1.3$,
$w_\text{capacity}{=}0.2$, $w_\text{balance}{=}0.1$.  This \emph{phase transition} is triggered
by the actual construction state, not by any distributional parameter.  It correctly reflects
the decision-making reality: when routes are nearly empty, placing high-demand customers in
capacity-rich routes prevents infeasibility regardless of the demand distribution; when routes
are nearly full, minimising incremental travel cost is paramount.

\paragraph{(2) Multi-objective coverage.}
The score balances five complementary signals---immediate travel cost, projected route length,
proximity, residual capacity, load balance---plus a route-size penalty.  No single distribution
shift can simultaneously mislead all signals.  For example, under a heavy-tailed demand
distribution the demand-scaled capacity term $w_\text{capacity}{\times}\texttt{residual}{\times}
\texttt{demand\_scale}$ strongly steers large-demand customers away from nearly-full routes,
preventing constraint violations; under a spatially clustered instance the proximity and
projected-route terms guide insertion toward the geographically nearest feasible route.

\paragraph{(3) Demand-adaptive capacity scaling.}
$\texttt{demand\_scale} = 1 + 0.5\,{\times}\,\max(0, d)$ amplifies the importance of
capacity whenever the arriving customer carries a large demand.  Rather than relying on
the expected demand from a training distribution, this adapts to the \emph{observed} demand
value at run time.  Under heavy-tailed demand distributions, where occasional very-large
customers appear, this prevents catastrophic insertions that leave a route with insufficient
capacity for subsequent customers.

\paragraph{(4) Balance score peaks at 50\% utilisation.}
The term $1 - 2|\texttt{current\_load} - 0.5|$ equals 1 at perfect half-load and $-1$ at
completely empty or fully loaded routes.  The 50\% target is distribution-agnostic: it
maximises the expected remaining capacity buffer regardless of whether future demands are
drawn from a uniform, Gaussian, or exponential distribution.  Heuristics that use a
distribution-specific optimal fill level degrade whenever the actual distribution shifts.

\paragraph{(5) Route-size penalty encourages spreading.}
Penalising large \texttt{route\_size} discourages concentrating many stops on a single
vehicle.  This implicit load-balancing across routes is beneficial across all distributions
because it maintains spare capacity on multiple routes simultaneously, reducing the chance of
total infeasibility when a burst of high-demand customers arrives---a concern particularly
acute under heavy-tailed or bursty demand distributions.

%% file: sections/appendix_cost_analysis.tex
\section{Computational Cost Analysis}
\label{app:cost}

This appendix quantifies the LLM API request budget, token consumption, and evaluation overhead of RAISE, and benchmarks these costs against five competing LLM-based AHD frameworks. All LLM-call counts follow the unified \emph{sample budget} convention used across the comparison (each budget unit $=$ one call to the LLM API resulting in one candidate heuristic program).

\subsection{RAISE Cost Breakdown}
\label{app:cost:RAISE}

Table~\ref{app:tab:cost_RAISE} lists the cost of a single RAISE run at the standard budget of $N_{\max} = 1000$ LLM samples, using a GPT-5-mini class model used in our experiments.

\begin{table}[htbp]
\centering
\caption{RAISE per-run cost breakdown ($N_{\max}{=}1000$, OBP primary configuration). Token counts are computed from the system design (Appendix~\ref{app:method_detail}); API costs use published list prices (GPT-5-mini: \$0.25/\$2.00 per 1M in/out tokens). Inner adversarial search uses CPU only (no LLM calls).}
\label{app:tab:cost_RAISE}
\small
\begin{tabular}{llr}
\toprule
\textbf{Component} & \textbf{Description} & \textbf{Quantity} \\
\midrule
\multicolumn{3}{l}{\textit{LLM API Calls (Outer Evolutionary Loop)}} \\
\quad Total budget $N_{\max}$ & Hard limit on LLM calls & 1,000 \\
\quad — \textsc{Create} & New heuristic & $\approx$20 \\
\quad — \textsc{E1} & Crossover & $\approx$245 \\
\quad — \textsc{E2} & Backbone extraction + Crossover & $\approx$245 \\
\quad — \textsc{M1} & Mutation & $\approx$245 \\
\quad — \textsc{M2} & Mutation & $\approx$245 \\
\midrule
\multicolumn{3}{l}{\textit{Token Consumption per LLM Call (approximate averages)}} \\
\quad \textsc{Create} & Task desc.\ + template & 1,200 in / 500 out \\
\quad \textsc{E1} & Task + 2 parents + instruction + template & 2,850 in / 650 out \\
\quad \textsc{E2} & Task + 2 parent + instruction + template & 3,050 in / 650 out \\
\quad \textsc{M1} & Task + parent + instruction + template & 1,850 in / 600 out \\
\quad \textsc{M2} & Task + parent + instruction + template & 1,650 in / 600 out \\
\quad \textbf{Weighted average} & & \textbf{2,200 in / 620 out} \\
\midrule
\multicolumn{3}{l}{\textit{Total Token Budget (1,000 calls)}} \\
\quad Input tokens & $1,000 \times 2,200$ & $\approx$2.2M \\
\quad Output tokens & $1,000 \times 620$ & $\approx$0.62M \\
\midrule
\multicolumn{3}{l}{\textit{Estimated API Cost}} \\
\quad GPT-5-mini & \$0.25/\$2.00 per 1M in/out & $\approx$\$\,1.79 \\
\midrule
\multicolumn{3}{l}{\textit{Inner Adversarial Search Overhead (CPU only; no LLM)}} \\
\quad instance refreshes & Every $\tau{=}5$ gens; 200 gens total & 40 \\
\quad Inner evaluations per refresh & $P_{\mathrm{in}}{=}8$ pop $\times$ $G_{\mathrm{in}}{=}4$ gens & 32 \\
\quad Re-scoring after refresh & $P_{max} = 10$ programs & 40 \\
\quad Outer evaluations per generation & $P_{max} = 10$ programs $\times$ 15 instances & 100 \\
\quad Total evaluations & $40{\times}(32{+}40) + 10{\times}15{\times}100 =  $ & $\approx$ 17,880\\
\midrule
\multicolumn{3}{l}{\textit{Wall-Clock Time (OBP, 20 parallel samplers/evaluators)}} \\
\quad Measured runtime &  & $\approx$2.0 h \\
\bottomrule
\end{tabular}
\end{table}

\subsection{Comparison with Existing Frameworks}
\label{app:cost:comparison}

Table~\ref{app:tab:cost_comparison} compares RAISE against EoH, ReEvo, PartEvo, MoH, and EoH-S at $N_{\max}{=}1000$ on online bin packing problem. 

\begin{table*}[htbp]
\centering
\caption{Per-run LLM cost comparison across AHD frameworks at $N_{\max}{=}1000$ samples. ${}^\dagger$ Estimated from published architecture. API cost uses GPT-5-mini: \$0.25/\$2.00 per 1M in/out tokens}
\label{app:tab:cost_comparison}
\small
\setlength{\tabcolsep}{5pt}
\begin{tabular}{lrrrr}
\toprule
\textbf{Method} &
\textbf{LLM Calls} &
\textbf{Total Tokens} &
\textbf{Est.\ Cost (mini)} &
\textbf{\#Instance Evals} \\
\midrule
EoH \citep{liu2024eoh}
  & 1,000 & $\approx$2.80M & $\approx$\$\,1.78 & $\approx$5,000 \\
ReEvo \citep{ye2024reevo}
  & 1,000 & $\approx$4.30M & $\approx$\$\,2.74 & $\approx$5,000 \\
PartEvo \citep{hu2025partition}
  & 1,000 & $\approx$4.51M & $\approx$\$\,2.89 & $\approx$5,000 \\
MoH \citep{shi2026generalizable}${}^\dagger$
  & 1,000 & $\approx$3.10M & $\approx$\$\,1.95 & $\approx$10,000 \\
EoH-S \citep{liu2026eohs}
  & 1,000 & $\approx$2.86M & $\approx$\$\,1.81 & $\approx$128,000 \\
\midrule
RAISE (ours)
  & 1,000 & $\approx$2.82M & $\approx$\$\,1.79 & $\approx$17,880 \\
\bottomrule
\end{tabular}
\end{table*}

\paragraph{Token overhead.}
All methods share the same 1,000-call budget. RAISE's total token cost ($\approx$2.82M) is comparable to EoH ($\approx$2.80M). ReEvo and PartEvo are substantially more expensive in tokens ($\approx$4.30M and $\approx$4.51M respectively) due to their multi-stage reflection and critique prompting strategies, which usually generate longer and more elaborate heuristic programs.

\paragraph{EoH-S incurs a hidden evaluation overhead.}
EoH-S has a token cost comparable to RAISE, but its instance evaluation count ($\approx$128,000) exceeds all other methods by over an order of magnitude. This arises from its \emph{post-hoc portfolio selection} step, in which the full heuristic population is evaluated across a diverse set of 128 training instances spanning multiple distributions to select the top-$K$ performers. Beyond the evaluation cost, this step requires \textbf{prior knowledge of deployment distributions}---often unavailable in practice. RAISE requires only nominal training-distribution instances, and delivers a single deployable heuristic rather than a portfolio.

\paragraph{MoH incurs dual-task evaluation overhead.}
MoH's $\approx$10,000 instance evaluations (double that of EoH) stem from its simultaneous training on two task variants---e.g., OBP under two item distributions---requiring every candidate heuristic to be scored on both variants each generation. This broadens coverage across the two training distributions but provides no adversarial robustness mechanism.

\paragraph{Inner adversarial search is essentially free at the API level.}
RAISE's inner worst instance search accounts for additional heuristic evaluations in CPU time only, which is independent of the number of deployment distributions, in contrast to EoH-S's evaluation overhead, which scales linearly with the size of its 128-instance training set.